\documentclass[12pt]{article}
\usepackage{amsmath,amsfonts}
\usepackage{algorithmic}
\usepackage{algorithm}
\usepackage{array}
\usepackage{subfig}
\usepackage{textcomp}
\usepackage{stfloats}
\usepackage{url}
\usepackage{verbatim}
\usepackage{graphicx}
\usepackage{cite}
\usepackage{caption}
\usepackage{booktabs}
\usepackage[T1]{fontenc}

\title{An unsupervised, open-source workflow for 2D and 3D building mapping from airborne LiDAR data}
\author{Hunsoo Song and Jinha Jung}

\begin{document}
\maketitle

% \markboth{IEEE JOURNAL OF SELECTED TOPICS IN APPLIED EARTH OBSERVATIONS AND REMOTE SENSING}%


\maketitle

\begin{abstract}
Despite the substantial demand for high-quality, large-area building maps, no established open-source workflow for generating 2D and 3D maps currently exists. This study introduces an automated, open-source workflow for large-scale 2D and 3D building mapping utilizing airborne LiDAR data. Uniquely, our workflow operates entirely unsupervised, eliminating the need for any training procedures. We have integrated a specifically tailored DTM generation algorithm into our workflow to prevent errors in complex urban landscapes, especially around highways and overpasses. Through fine rasterization of LiDAR point clouds, we've enhanced building-tree differentiation, reduced errors near water bodies, and augmented computational efficiency by introducing a new planarity calculation. Our workflow offers a practical and scalable solution for the mass production of rasterized 2D and 3D building maps from raw airborne LiDAR data. Also, we elaborate on the influence of parameters and potential error sources to provide users with practical guidance. Our method's robustness has been rigorously optimized and tested using an extensive dataset (> 550 {km\textsuperscript{2}}), and further validated through comparison with deep learning-based and hand-digitized products. Notably, through these unparalleled, large-scale comparisons, we offer a valuable analysis of large-scale building maps generated via different methodologies, providing insightful evaluations of the effectiveness of each approach. We anticipate that our highly scalable building mapping workflow will facilitate the production of reliable 2D and 3D building maps, fostering advances in large-scale urban analysis. The code will be released upon publication.
\end{abstract}

\section{Introduction} \label{Introduction}
\subsection{Current State of Large-area 2D Building Mapping}
Buildings are key structures in which numerous human activities unfold. They offer invaluable insights into human practices and the subsequent environmental impacts \cite{zhu2019understanding}. Building maps, especially those derived from remote sensing imagery, are crucial to numerous fields, including disaster management, urban ecology, smart city planning, population estimation, and humanitarian aid \cite{ghaffarian2019post,hong2019temporal,herbert2015comparison,wu2005population,herfort2021evolution, harig2021automatic}. However, the inherent uncertainties and errors in building maps can mislead studies reliant on them. Thus, the remote sensing community is dedicated to enhancing the quality of such maps.

Publicly available building maps typically provide only 2D information due to dependence on optical imagery. For instance, OpenStreetMap (OSM) provides extensive spatial coverage but lacks 3D information and suffers from inconsistencies in completeness, accuracy, and data vintage \cite{hecht2013measuring, vargas2020openstreetmap}. While authoritative maps, often regarded as ground truth \cite{dorn2015quality,jokar2015quality}, provide better accuracy but still lacks 3D information, in most cases, and have limited spatial coverage due to high production costs. This results in a trade-off between the spatial coverage and accuracy of publicly accessible building maps.

As it stands, large-scale building maps spanning metropolitan areas - instrumental for multi-city-scale analyses - are predominantly 2D building maps created based on optical imagery. 
The rise of deep learning and several contests have facilitated large-area 2D building mapping. A notable example is the SpaceNet challenge \cite{van2018spacenet}. In this challenge, deep learning-based methods have shown the highest accuracy. Since this challenge, deep learning-based methods have been dominating literature for the automatic generation of building footprints \cite{huang2023building, ma2019deep, zhu2017deep}. However, their accuracies vary considerably depending on the region and the training condition \cite{yang2018building}. Despite ongoing efforts like transfer learning and domain adaptation \cite{maggiori2016convolutional,deng2019large,makkar2021adversarial, dias2022model}, creating universally applicable models remains a challenge.

One milestone in a large-area building mapping was Microsoft Building Footprints \cite{MS_BUILDING}. Microsoft Building Footprints released the largest building footprints that were generated by a machine. A deep learning model called EfficientNet \cite{tan2019efficientnet} was trained with millions of building labels and their corresponding satellite images, and the outputs of the model were refined based on a polygonization algorithm to produce the final building footprint maps. Microsoft Building Footprints is a significant accomplishment in that it provides the first continental-level open-building maps generated by algorithms. However, similar to OpenStreetMap, it has errors, and its quality is not consistent as diverse optical images with different conditions were used for the mapping \cite{heris2020rasterized}. 

Given that the current accuracy of deep learning-based mapping using optical imagery is not sufficient to supplant authoritative maps - even with millions of satellite images and corresponding labels - it suggests the need for another breakthrough to bridge the accuracy gap. This is particularly challenging in regions where available training labels and images are limited and the building appearances significantly vary \cite{williams2019mapping}.

\subsection{Towards Large-area 3D Building Mapping}
Beyond the accuracy of 2D building footprint maps, it is important to remember that buildings are inherently 3D entities. With cities becoming denser and taller, a 3D perspective becomes crucial for understanding urban developments and their environmental impacts \cite{li2020continental,herbert2015comparison}. For instance, 3D building maps can deepen the understanding of urban and climate studies \cite{park2021impacts}, aid disaster management \cite{macchione2019moving}, and improve the accuracy of population estimation \cite{wang2016fine,biljecki2016population}. They not only enrich information for downstream analyses \cite{biljecki2016variants} but also open new research domains, including urban air mobility navigation \cite{kim2021regionalization}, digital twin simulations \cite{lehner2020digital}, and various smart city applications \cite{han2022utilising,zhou2019community,kamra2022lightweight}.

Airborne laser scanning (ALS) is recognized as the most efficient method for capturing detailed 3D building information, given sufficient point density. Despite the expense, ALS offers a viable alternative to optical or SAR sensors \cite{qin2019critical, huang2022evaluation, li2020continental,dominguez2019back}, which often rely on specific conditions and can struggle to yield comprehensive 3D data. In particular, if detailed 3D building information is required, or if LiDAR can provide a more accurate 2D building map, ALS could be a better option for both 2D and 3D building mappings, considering the cost and scalability of optical imagery-based building mapping \cite{yang2018building, van2018spacenet}. Consequently, if a robust workflow for 2D and 3D building mapping from ALS is established, LiDAR-based building mapping could be more appealing than other methods in terms of both accuracy and cost.

Over the past several decades, a plethora of algorithms for automatic 3D building mapping using airborne LiDAR data have emerged. This includes deep learning-based techniques and those fostered by competitions, such as those initiated by the International Society for Photogrammetry and Remote Sensing (ISPRS) community \cite{rottensteiner2012isprs, rottensteiner2014results}. While these algorithms have shown their potential in mapping small datasets (< 100 buildings), their scalability and efficiency for large-area mapping remain uncertain due to the limited diversity and spatial coverage of their study areas. This leads to concerns about their performance in diverse landscapes \cite{tuia2016domain,dias2022model}. Moreover, the specifics of implementing these algorithms are often undisclosed, hindering their widespread application and further refinement.

Recent research efforts have been directed towards the extraction of intricate building structures, such as roofs and facades, aiming for the realistic ``3D modeling'' of buildings \cite{wang2023automatic, liu2023roof, wang2023reconstruction, huang2022city3d, lewandowicz20223d, li2022recursive, bizjak2023novel, dey2023machine, li2022ransac, zhang2021optimal, li2022point2roof, tarsha2022automatic}. These techniques have enhanced the granularity of building models and would be apt for projects requiring detailed representation in CityGML \cite{groger2012citygml} or Building Information Modeling (BIM) \cite{volk2014building}. However, their emphasis on details often hinders the scalability and efficiency required for large-scale mapping. Additionally, these methods typically operate under the assumption that buildings primarily consist of planar or rectangular features, thus limiting their generalizability \cite{kamra2022lightweight}.

This paper presents an end-to-end, open-source workflow aimed at enabling the mass production of rasterized 2D and 3D building maps. Our research prioritizes efficiency and scalability, setting it apart from prior studies. This approach ensures our method is well-suited for large-scale mapping across diverse landscapes, encompassing not only complex urban areas intertwined with various urban structures such as overpasses and highways but also rural and densely forested regions. Our workflow, requiring only discrete point cloud data from typical topography ALS, outputs reliable 2D and 3D building maps with no need for parameter tuning, delivering building footprints of superior accuracy compared to those in Microsoft Building Footprints.

Our open-source workflow is designed to operate in an unsupervised manner, leveraging a simple yet robust assumption rooted in the physical properties of buildings - their laser-impermeability and relative smoothness as ground-standing objects. This focus enhances scalability and mitigates unexpected errors, a critical aspect in large-area mapping. Furthermore, the workflow integrates terrain modeling, a critical yet often overlooked component in previous small-scale studies. 

The workflow's effectiveness has been honed through rigorous tests in various areas and comprehensive evaluations against authoritative maps and Microsoft Building Footprints over an expanse of 550 {km\textsuperscript{2}} urban areas. We also elaborate on potential error sources in LiDAR-based building maps to guide potential users and offer comprehensive evaluations against deep learning-based products and hand-digitized maps. In sum, our open-source workflow is primed to expedite the mass production of large-area 3D building maps, yielding valuable data for urban and environmental studies on a city-wide or national scale.

The main contributions can be summarized as follows.

\begin{itemize}
  \item We present an open-source workflow that can facilitate the mass production of rasterized 2D and 3D building maps for large-area using airborne LiDAR data.
  
  \item Our carefully engineered workflow can produce more accurate building maps with greater accuracy than Microsoft Building Footprints, delivering robust performance across varied landscapes without the need for parameter tuning.
  
  % \item The impact of parameter selection and potential sources of error are detailed to assist potential users.
  
  \item Our study provides a comprehensive comparison of building maps derived from three different methods: LiDAR, deep learning with optical imagery (specifically, Microsoft Building Footprints), and hand-digitized building maps.
  
\end{itemize}

The remainder of this paper is organized as follows. Section II describes the proposed workflow and illustrates the procedure for the optimization. Section III elaborates on experimental results and analyzes the errors. Section IV discusses the impact of parameter selections and the limitations of our workflow. Section V concludes the study.

\section{Methodology} \label{s2.}
\label{sec:headings}

%\lipsum[4] See Section \ref{sec:headings}.

\subsection{Overview} \label{s2.1.}

\begin{figure*}
	\centering
	\includegraphics[width=1\textwidth]{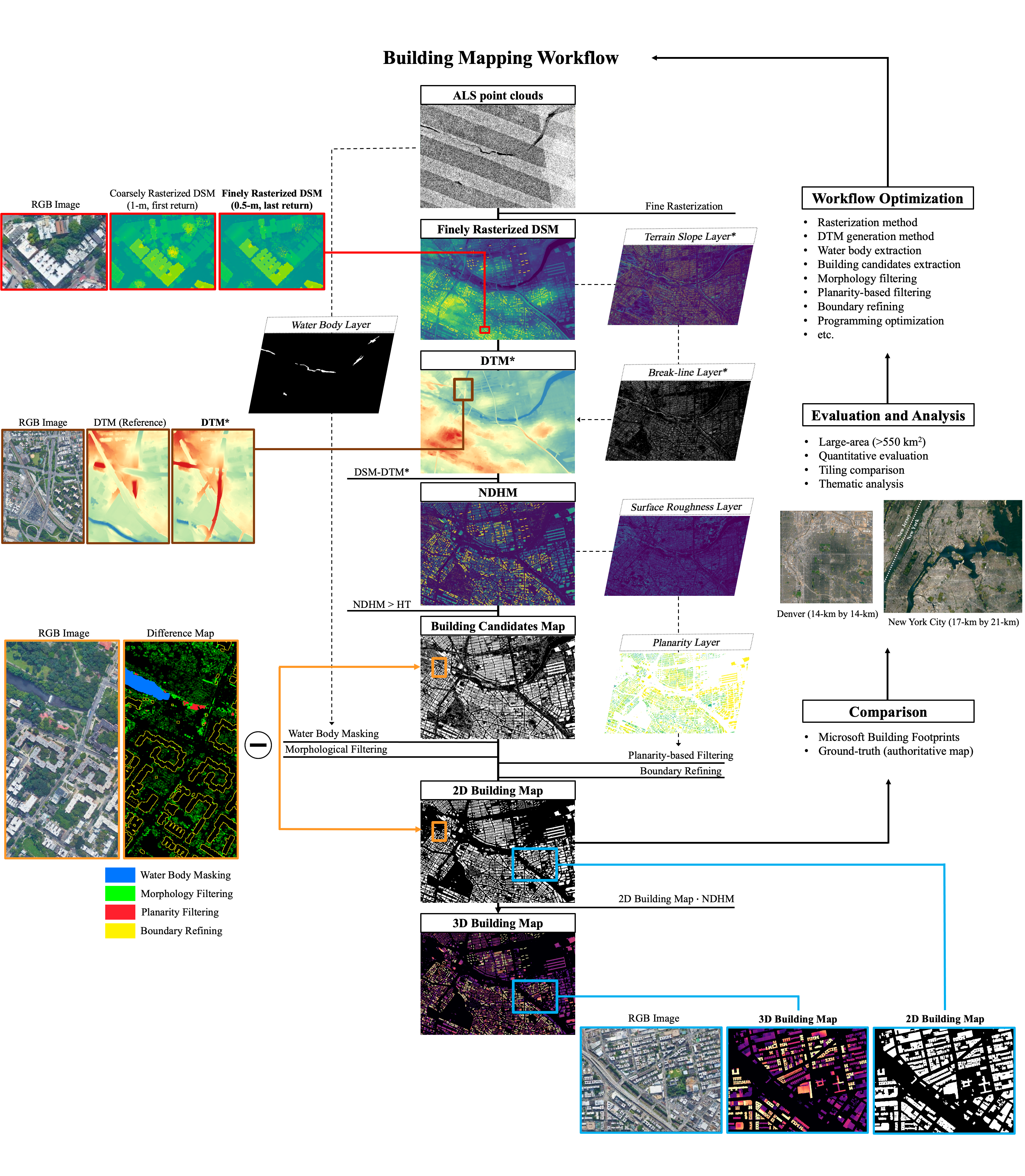}
	\captionsetup{justification=centering}
	\caption{Overview of the methodology}
	\label{fig:fig0}
\end{figure*}

In this Section, we detailed the proposed workflow for generating 2D and 3D building maps. The proposed workflow generates rasterized 2D and 3D building maps from the raw ALS point clouds. The workflow operates in a simple but robust rule-based approach by exploiting the physical properties of buildings, which are ground-standing and have laser-impermeable, relatively smooth surfaces.
The workflow has been optimized by iterative algorithm developments and evaluations. For evaluation, the output from the workflow was compared to Microsoft Building Footprints and ground-truth (authoritative map). Specifically, two large cities, Denver and New York City, covering more than 550 {km\textsuperscript{2}} in total were used for the evaluation. In addition to conventional quantitative evaluation, a tiling comparison method that evaluates maps based on ranking after tiling them into small tiles was introduced to compare different large-area building maps effectively. Also, thematic error analyses were conducted based on diverse criteria. The following subsections will illustrate the proposed workflow and detailed descriptions of evaluation methods for optimizing the workflow. Figure \ref{fig:fig0} provides an overview of the methodology section, summarizing the proposed building mapping workflow and the procedures for the optimization.% The area shown in this figure is a sample of 3-km by 4.5-km from the Bronx, New York. The laser scanning was conducted with a Riegl Q-1560 LiDAR system with the point density of approximately 7-points/{m\textsuperscript{2}}.

% Comment: maybe we can add a figure that shows the overall workflow of the proposed algorithm here? That way, readers will have a better idea on what will follow after? We can add some additional text explaining the figure as well. 

\subsection{The proposed workflow} \label{s2.2.}
\subsubsection{From ALS point clouds to Finely Rasterized DSM} \label{s2.2.1.}
%\noindent\emph{PART I: from ALS point clouds to Finely Rasterized DSM}
%\textbf{PART I: From ALS point clouds to Finely Rasterized DSM}

Our building mapping workflow starts from the raw point cloud collected from a typical ALS system. The point cloud represents 3D coordinates of data points that are measured from the airborne LiDAR sensor. Since the observed data point is a subset of the points on the earth's surface, it necessarily has a limitation in representing the earth perfectly. The point observation can be too sparse, and the point spacing inevitably varies due to many factors such as flight configuration and the laser reflectivity of objects on the ground. To handle this irregularly spaced raw data, the raw point cloud is commonly transformed into a gridded format, called the digital surface model (DSM). During the transformation, depending on the ground sampling distance (GSD) of the DSM, the void grid where the LiDAR point does not register can occur. 

One typical way to avoid this void grid is to use a coarser GSD so that majority of grids will have at least a LiDAR point registered and to take a representative height value among points in the grid \cite{hyyppa2001segmentation,jung2014framework,maltezos2018building,oh2022high}. Studies using both LiDAR and optical imagery usually generated DSM that has the same resolution of corresponding optical imagery regardless of the point density \cite{awrangjeb2012building,huang2019automatic,chen2020automatic}. Occasionally, how to create and rasterize a DSM from LiDAR has not been explicitly mentioned in the previous literature \cite{zhao2016extracting, yan2017hierarchical, yuan2021multiscale} because it is often not considered a critical issue, or even if it is mentioned, they simply provide the software used to create the DSM (e.g. LasTools, ArcGIS)\cite{hosseinpour2022cmgfnet, ojogbane2021automated}. However, creating a DSM of coarse grid (``coarsely rasterized DSM'') has two problems. First, it can cause data loss. Second, it can make the differentiation between buildings and trees more difficult as it may lose the feature that can be obtained from the characteristics of laser penetration.

Instead of coarse rasterization, our workflow uses a ``fine rasterization'' for generating DSM. Fine rasterization is a method that projects a point cloud into finely rasterized grids and interpolates void grids to reduce cases where multiple LiDAR points occupy a common grid. Before the interpolation, fine rasterization will obviously result in more void grids than coarse rasterization and will require interpolation of non-observed values. However, it can prevent data loss that occurs when multiple LiDAR points are registered to the same grid. Also, the interpolation can reasonably extend the observation unless point density is severely irregular \cite{morgan2002interpolation}. Specifically, in our workflow, the LiDAR point whose elevation is the lowest (the last return of LiDAR points) was chosen for the DSM value when multiple points occupy the same grid in fine rasterization. This increases the chance of collecting the penetrated laser points under the trees, and subsequently, penetrated laser points make the differentiation between buildings and trees easier. For the interpolation, the nearest interpolation is selected as we found it prevents distortion of height values at building boundaries.

\begin{figure}
	\centering
	\includegraphics[width=3.2in]{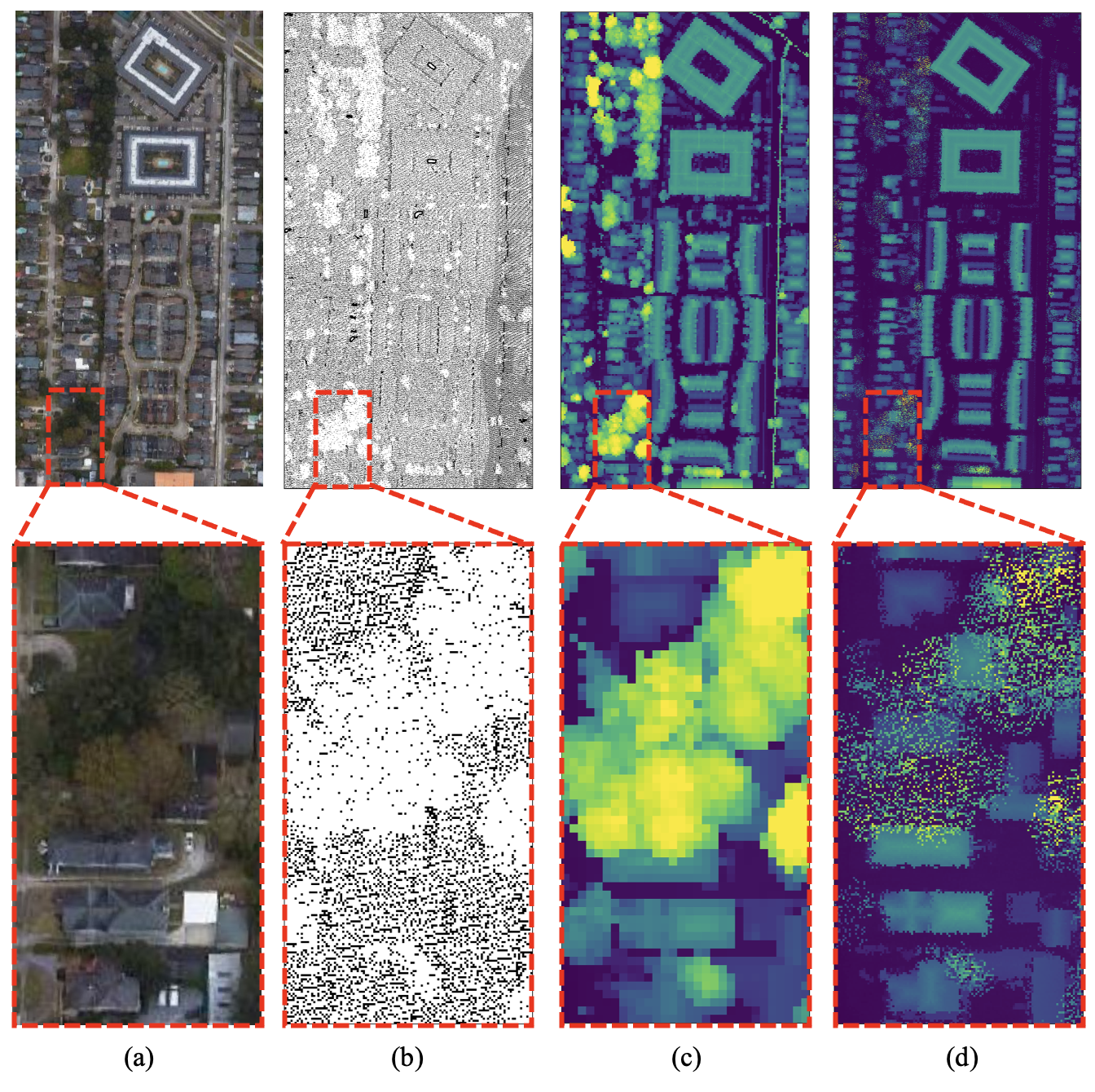}
 	%\captionsetup{justification=centering}
	\caption{Comparison of two DSM rasterization methods: (a) RGB image, (b) LiDAR occupancy map, (c) coarsely rasterized DSM, (d) finely rasterized DSM.}
	\label{fig:fig1}
\end{figure}

Figure \ref{fig:fig1} emphasizes how DSM can be different depending on the rasterization method. The area shown in this figure is a sample from New Orleans, Louisiana, US, where overhanging dense trees often cover residential buildings. The area of the top figure is 0.65 km by 0.30 km, and the bottom figure shows the zoomed-in area of the top figure. The laser scanning was conducted with Riegl LMS-Q680i, and the point density is approximately 4-points/{m\textsuperscript{2}}. The RGB satellite image from Google Earth (Figure \ref{fig:fig1}(a)) is provided for reference. Figure \ref{fig:fig1}(b) shows LiDAR point occupancy, which displays the grid occupied by LiDAR points in white, otherwise black. The black grids in LiDAR point occupancy represents void grids of DSM before the interpolation. The two different DSMs created with coarse rasterization and fine rasterization are shown in Figure \ref{fig:fig1}(c-d), respectively. The coarsely rasterized DSM took the highest elevation when multiple points coexist in the same grid while the finely rasterized DSM took the lowest elevation. The GSD of the coarsely rasterized DSM and the finely rasterized DSM are 2-meter and 0.5-meter, respectively. As shown, coarse rasterization made it difficult to distinguish between buildings and trees. On the contrary, fine rasterization better retain buildings shapes while trees are illustrated as scattered points, which stands out the difference between buildings and trees becomes clearer. In other words, fine rasterization dramatizes the difference in penetration properties of buildings and trees.\newline

%\paragraph{Paragraph}
%\lipsum[7]

\subsubsection{From Finely Rasterized DSM to DTM* and NDHM}
%\noindent\emph{PART II: from Finely Rasterized DSM to DTM* and NDHM}

A building can be characterized as an object that stands on the ground, which means buildings are relatively tall and have a discrete height difference from the nearby ground along with their boundaries. Therefore, simply calculating the relative height above the nearby ground and masking with a certain height elevation can effectively extract building candidates. For this, the generation of the digital terrain model (DTM), the map describing the ground elevation, must precede to generate a normalized digital height model (NDHM). The NDHM is the surface model that represents the height above the ground, and it can be produced by subtracting DTM from DSM. 

\begin{figure*}
	\centering
	\includegraphics[width=6.25in]{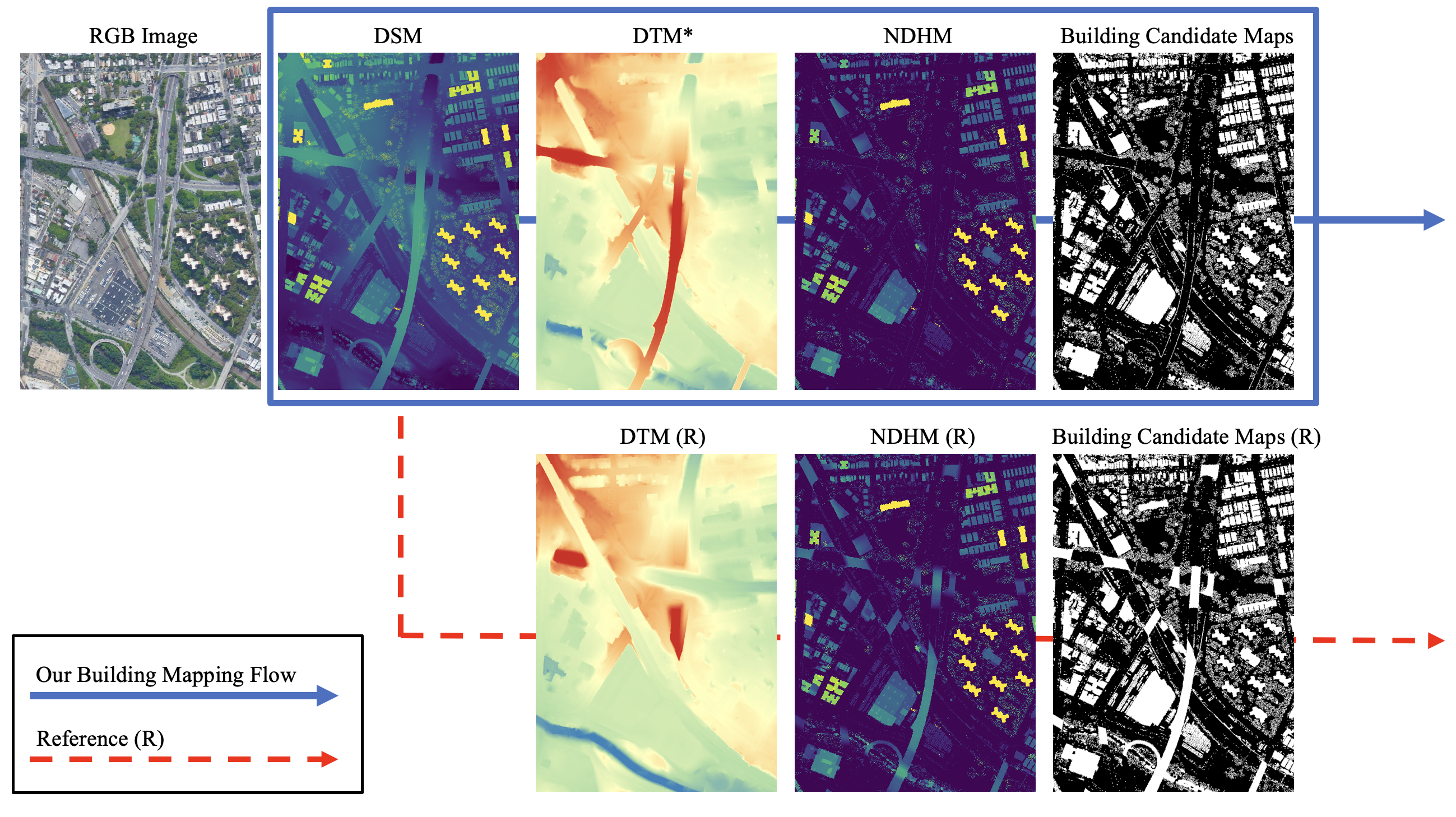}
 	%\captionsetup{justification=centering}
	\caption{Comparison of DTM* and a reference DTM and their respective results. DTM* can prevent overpasses from becoming building candidates.}
	\label{fig:fig1.5}
\end{figure*}

Our workflow incorporates a method for generating DTM (``DTM*'')\cite{song2022dtm}. This method is preferred due to its computational efficiency and ability to maintain detailed object boundaries, aspects that are critical for 3D building mapping. DTM* identifies an object as a region entirely enclosed by steep slopes. The process initiates with the creation of a break-line map that delineates steep slopes. Following this, a connected component algorithm is employed, filtering regions fully surrounded by steep slopes as objects while ensuring all ground sections are interconnected smoothly. This unique object-wise filtering performed by DTM*, which outlines the boundaries of ground-standing objects using the same DSM, enhances computational efficiency and helps to avert common errors often associated with large buildings and their boundary definitions.
Another crucial reason for adopting DTM* is that it considers bridges and overpasses as ground, unlike typical DTM generation methods \cite{sithole2003report, meng2010ground}. This property is important because otherwise, the ground beneath the bridges and overpasses will become a DTM, resulting in errors in extracting the bridges and overpasses as building candidates in the subsequent process of the workflow. Figure \ref{fig:fig1.5} compares DTM* and a reference DTM (``DTM(R)'') and shows their respective results. The reference DTM was acquired from the U.S. Geological Survey. Unlike the reference DTM, DTM* treats overpass as ground, making them different from buildings in the NDHM and Building Candidate Maps in subsequent steps of the workflow. %In addition, the water body extraction of DTM* was modified and integrated into our workflow so that areas detected as water bodies were masked out in subsequent building candidate extraction. Detailed procedures for building candidate extraction and water body masking are provided in the latter paragraphs.\newline

\subsubsection{From NDHM to Building Candidates Map, 2D and 3D Building Maps} \label{s2.2.3.}
%\noindent\emph{PART III: from NDHM to Building Candidates Map, 2D and 3D Building Maps}

% An abbreviated procedure from after the finely rasterized NDHM generation is visualized in Figure \ref{fig:fig2} along with a RGB satellite image from Google Earth Figure \ref{fig:fig2}(a).
Building Candidates Map is generated by applying a mask with a certain Height Threshold (HT) to the NDHM. Building Candidates Map will represent all objects that are relatively taller than the nearby ground as a binary format, and the objects become building candidates. As overpasses were treated as ground, building candidates would be either buildings, non-building small objects, or noises. To extract only buildings out of all building candidates, the workflow performs four series of operations: (1) water body masking, (2) morphological filtering, (3) planarity-based filtering, (4) boundary refining. The result of these operations is 2D Building Map. 

% Lastly, our workflow refines building boundaries by applying a dilation kernel of size K3 (Figure \ref{fig:fig2}(i)). The dilation is to refine noisy building boundaries and to restore the underestimated building area because LiDAR can underestimate building boundaries. This underestimation is detailed in Section IV-B. Finally, the workflow generates a 3D building map by extracting building pixels of the 2D building map from the NDHM.

\begin{figure*}
	\centering
	\includegraphics[width=1\textwidth]{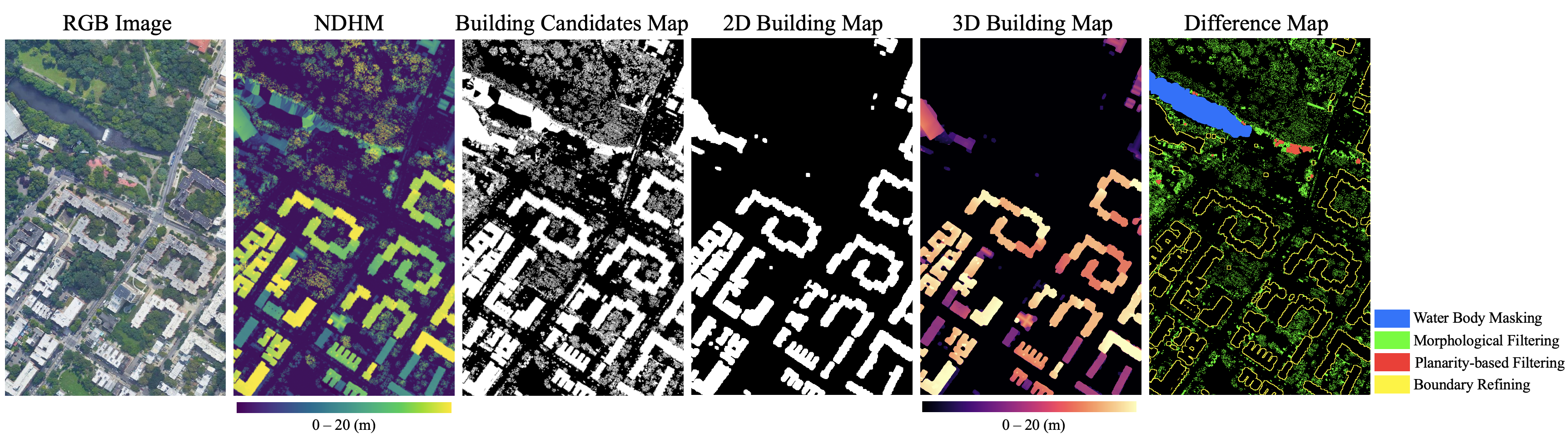}
 	%\captionsetup{justification=centering}
	\caption{From NDHM to Building Candidates Map, 2D and 3D Building Maps. The color in Difference Map indicates which operation caused the difference between Building Candidates Map and 2D Building Map.}
	\label{fig:fig2}
\end{figure*}

The first operation is the water body masking. The general idea of water extraction is similar to \cite{hofle2009water, song2022dtm} in that it takes advantage of the fact that the point density above water is much lower than others. However, the point density also can be low due to occlusion by tall objects such as buildings or noise from the LiDAR points themselves. In addition, as the water body contains sparse and noisy LiDAR points, it often creates large planar objects near water, which are likely to be extracted as buildings in the subsequent process. For example, when trees are surrounded by water pixels, they are likely to be extracted as buildings because empty grids caused by water are interpolated with the elevation values of nearby trees. Our workflow begins by classifying surface water bodies based on local point density, specifically by counting the number of LiDAR points in a sliding window and designating the center pixel as surface water if the number of pixels within the window is significantly smaller than the average, based on the binomial distribution. By default, we use a 9 by 9 window in 0.5-m resolution DSM and classify it as a water class if it is 2 sigma or more below the average point density. We further refine the process with two additional rules: first, we exclude smaller water bodies (those less than 1,000{m\textsuperscript{2}}) from masking, and second, we apply a 5-meter water buffer to the remaining large water bodies. These additional steps mitigate the risk of erroneous building extraction due to noise around the water and prevent buildings from being incorrectly masked out due to occlusion. %The blue area in Difference Map of Figure \ref{fig:fig2} shows the area masked-out by the water body layer.

%%%%%%%%%%%%%%%%%%%%%%%%%%%%%%%%%%%%%%%%%%%%%%%%%%%%%%%%%%%%%%%%%%%%%%%%%%%%%%%%%%%%%%%%%%%%%%%%
The second operation is morphological filtering. One distinctive difference between buildings and other building candidates is that buildings are generally larger than several tens of square meters. Particularly for trees, as LiDAR can penetrate through trees and observe the ground under the tree, trees are represented like salt and pepper noise in Building Candidates Map. On the other hand, LiDAR cannot penetrate through the building roof as its surface is solid (laser-impermeable) and flat that consistently representing higher elevation than its surrounding ground. The different characteristics had become especially distinct with fine rasterization as shown in Figure \ref{fig:fig1}. To exploit the difference, our workflow applies morphological filters of erosion and dilation consecutively on Building Candidates Map. The erosion filter removes small objects by eroding pixels, and the dilation filter restores the eroded pixels only if they remain after the erosion. %This morphological filtering removes most of non-building objects as depicted in Figure \ref{fig:fig2}. 
As building mapping is essentially a binary classification, a trade-off between omission and commission errors necessarily exists. When applying the morphological filter, the kernel size (K1) must be determined appropriately to consider the trade-off. If K1 increases, the commission error decreases as non-building small objects can be removed more aggressively after the erosion. On the other hand, the omission error increases as small buildings may be removed during the erosion process. Conversely, a smaller kernel will increase commission errors while decreasing omission errors. The trade-off caused by the choice of K1 is analyzed in Section IV-B.

The third operation is planarity-based filtering. One common feature of buildings in NDHM is that their local height variations are low. Extracting buildings based on their textual feature had become one of the standard approaches to extracting buildings. To quantify the local height variation, several approaches have been used, namely, co-occurrence matrix-based \cite{liu2013automatic, du2017automatic}, eigenvalue-based \cite{niemeyer2014contextual, rottensteiner2003automatic}, Entrophy-based \cite{maltezos2018building}. Here, we followed a similar strategy but used a new, computationally efficient way to calculate the local height variation. First, our workflow rounds the NDHM to have integer values and counts the number of unique integers within a square kernel (K2) over the NDHM. Then, the count of the number of unique integers represents the roughness of each pixel. This process generates a surface roughness layer (shown in Figure \ref{fig:fig0}). Each pixel in the surface roughness layer represents the number of unique integer values within a square window of size K2. Subsequently, a pixel is categorized as `planar' if its roughness value is less than a predefined Roughness Threshold (RT). For instance, consider a case where K2 is 5 and RT is 4. If the number of unique integer values in a 5 by 5 window of a rounded NDHM is less than 4, the algorithm identifies the center pixel of the window as planar.
The algorithm then calculates the proportion of planar pixels for each remaining building candidate, a measure referred to as `planarity'. For example, if a building candidate comprises 100 pixels and 20 of them are planar, the planarity of the candidate is 0.2. The planarity layer in Figure \ref{fig:fig0} illustrates planarity values for all building candidates.
Assuming that building roofs have a planar surface compared to non-building objects, we employ a planarity-based filtering algorithm to exclude objects with planarity values less than a specified ratio, termed the Dense Tree (DT) value. This value is employed as a descriptor to differentiate dense trees from building candidates.
A more detailed description of DT is provided in Section IV-B. 
%The red area in Difference Map of Figure \ref{fig:fig2} shows the area masked-out by the planarity. 

The fourth operation is boundary refining. Our workflow refines building boundaries by simply applying a dilation kernel of size K3, unlike other approaches that commonly use both erosion and dilation \cite{yu2010automated}. This is to mitigate the deformation of building boundaries and to restore the underestimated building area because LiDAR can underestimate building boundaries. This deformation and underestimation are detailed in Section IV-B. Since the main purpose of our workflow is an efficient, well-generalizable large-area building mapping rather than detailed modeling of buildings, it does not use complex boundary refining methods \cite{dos2019extraction,dos2022weighted, kamra2022lightweight} that often require constraints on the building shape.

Finally, the workflow generates a 3D building map by extracting building pixels of the 2D building map from the NDHM. 
Figure \ref{fig:fig2} shows a part of the workflow from NDHM to 2D and 3D Building Maps. ``Difference Map'' represents the difference between Building Candidates Map and the 2D Building Map. The color in Difference Map indicates which operation caused that difference. Our workflow generates building maps of 0.5-meter resolution as default and has been optimized through extensive experimental results. The followings describe default parameter values. HT was set as 1.5 meters. The kernel size (K1) of morphological filters was set as 7 (a 7 by 7 pixels window). A 5 by 5 pixels window was used for K2. RT and DT were set as 4 and 0.1, respectively. K3 was set as a 5 by 5 pixels window by default. Although the best parameter combination must be different depending on the specific area, the default parameter values were found to be robust through an optimization process.

\subsection{Optimization and Evaluation} \label{s2.3.}

\begin{figure}
	\centering
	\includegraphics[width=1.4in]{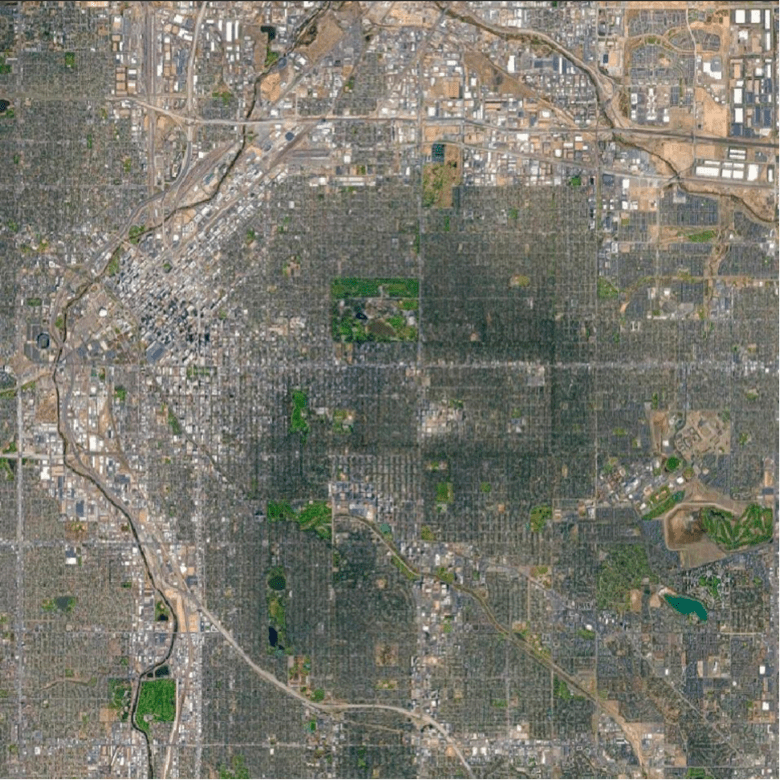}
	\caption{Study area of Denver (14-km by 14-km)}
	\label{fig:fig3}
\end{figure}

\begin{figure}[t]
	\centering
	\includegraphics[width=2.1in]{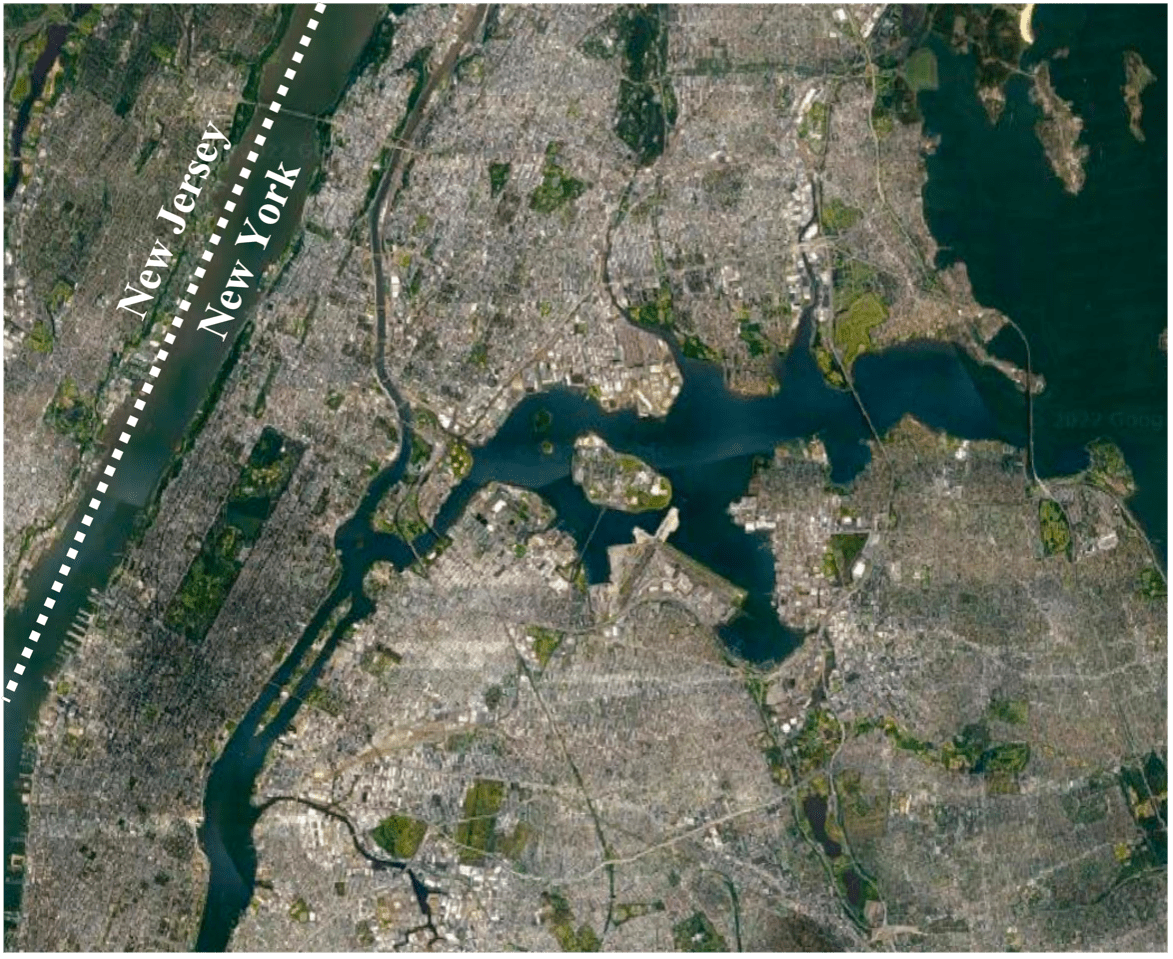}
	\caption{Study area of New York City (17-km by 21-km, New Jersey is excluded)}
	\label{fig:fig4}
\end{figure}

%To optimize the workflow based on the quantitative results, w
We examined the workflow using large urban area datasets of Denver, Colorado, US (196 {km\textsuperscript{2}}) and New York City, New York, US (357 {km\textsuperscript{2}}) where both authoritative building maps and Microsoft Building Footprints are available. With these datasets, algorithms and parameters of the workflow have been optimized by repetitive performance evaluations and developments. The aerial RGB imagery of two study areas are shown in Figure \ref{fig:fig3} and Figure \ref{fig:fig4}, respectively. The aerial RGB imagery are from the U.S. Department of Agriculture’s (USDA) National Agriculture Imagery Program (NAIP)'s orthoimagery. Laser scanning of the Denver area was performed with the Leica TerrainMapper sensor between May and September 2020. For New York City, the Leica ALS70 sensor was used between March and April 2014. All LiDAR data used in this study are from the U.S. Geological Survey’s 3D Elevation Program (3DEP). The point densities of the Denver and New York City datasets are approximately 4-points/{m\textsuperscript{2}} and 5-points/{m\textsuperscript{2}}, respectively. Then, building maps generated by our workflow (``LiDAR building map'') are compared to both authoritative building maps (``ground-truth'') and Microsoft Building Footprints. All building maps were evaluated as raster with 0.5-meter resolution.%All building maps were rasterized to have 0.5-meter resolution for the quantitative evaluation.

Our study area is over 550 {km\textsuperscript{2}}. Conventionally, building extraction methods were evaluated by providing several averaged quantitative metrics such as intersection over union (IoU), precision, recall, and F1-score of entire study areas. However, as our study area is large, simply investigating averaged metrics of the entire area was not sufficient to optimize parameters and evaluate the workflow in detail. Therefore, in addition to the conventional quantitative metrics, we introduced a tiling comparison method to explore the weaknesses and strengths of the proposed workflow in detail. The tiling comparison method is a method that compares different maps by dividing them into small tiles. The tiling comparison method has the advantage of being able to effectively show distinct differences among large-area maps by providing a ranking of the differences. It also enables effective comparative evaluations of different maps whose characteristics may vary from region to region.

To be specific, we tiled each study area so that the area of each tile has an area of 0.5 km by 0.5 km. As a result, Denver and New York City were tiled to have 784 tiles and 1428 tiles, respectively. With these tiles, we calculated IoUs between the LiDAR building map and Microsoft Building Footprints for every tile. Then, we ranked the IoUs to find tiles having significant differences between LiDAR building map and Microsoft Building Footprints. Based on the rank of the difference, we qualitatively compared their performances. Also, the IoU of LiDAR building map and the IoU of Microsoft Building Footprints were also calculated by comparing to the ground-truth, respectively. Lastly, we analyzed the errors of generated building maps according to the building areas. Section III provides the result with the default parameters.

% \begin{figure*}[hbp]
% 	\centering
% 	\includegraphics[width=0.6\textwidth]{figures/fig3.png}
% 	\caption{Study area of Denver (14-km by 14-km)}
% 	\label{fig:fig3}
% \end{figure*}

\section{Results}\label{s3.}
\subsection{Results of the Denver dataset} \label{s3.1.}
Three building maps: LiDAR building map,  Microsoft Building Footprints, and ground-truth (authoritative map) were compared. The LiDAR building map is the result of the workflow with default parameter values described in Section II-B. One thing to note is that there are time discrepancies in each building map. The vintage of ground-truth for Denver is 2018, while LiDAR building map was generated with LiDAR data for 2020, and Microsoft Building Footprints were generated with optical images of 2018-2019.\newline

\subsubsection{Conventional quantitative results} \label{s3.1.1.}
LiDAR building map and Microsoft Building Footprints of the entire study area (196 {km\textsuperscript{2}}) of Denver were evaluated in terms of IoU, precision, recall, and F1-score (Table~\ref{tab:table1}). As a result, our workflow outperformed Microsoft Building Footprints in all metrics. IoU and recall were particularly higher than that of Microsoft Building Footprints. Considering the LiDAR building map was generated in a fully unsupervised way with a single default parameter set, the result shows that our workflow can produce a building map more accurately than Microsoft’s deep learning-based method as long as a decent quality of ALS data is available.\newline

\begin{table}[ht]
	\caption{Conventional quantitative results of the Denver dataset}
	\centering
	\begin{tabular}{ccccc}
%	\begin{tabular}{lllll}	
		\toprule
% 		\multicolumn{2}{c}{Part}                   \\
% 		\cmidrule(r){1-2}
		& IoU     & Precision & Recall & F1-score \\
		\midrule
		Our workflow & \textbf{81.8} &  \textbf{91.2} & \textbf{88.8} & \textbf{90.0}\\
		\midrule
		Microsoft's & 77.3 &  90.4 & 84.2 & 87.2\\
%		LiDAR building map & 81.8 & 91.2 & 88.8 & 90.0\\
		\bottomrule
		
	\end{tabular}
	\label{tab:table1}
\end{table}
% \begin{table}[ht]
% 	\caption{Conventional quantitative results of the Denver dataset}
% 	\centering
% 	\begin{tabular}{ccccc}
% %	\begin{tabular}{lllll}	
% 		\toprule
% % 		\multicolumn{2}{c}{Part}                   \\
% % 		\cmidrule(r){1-2}
% 		& IoU     & Precision & Recall & F1-score \\
% 		\midrule
% 		LiDAR building map & \textbf{81.8} &  \textbf{91.2} & \textbf{88.8} & \textbf{90.0}\\
% 		\midrule
% 		Microsoft Building Footprints & 77.3 &  90.4 & 84.2 & 87.2\\
% %		LiDAR building map & 81.8 & 91.2 & 88.8 & 90.0\\
% 		\bottomrule
		
% 	\end{tabular}
% 	\label{tab:table1}
% \end{table}

\subsubsection{Tiling comparison} \label{s3.1.2.}
\begin{figure*}
	\centering
	\includegraphics[width=5.5in]{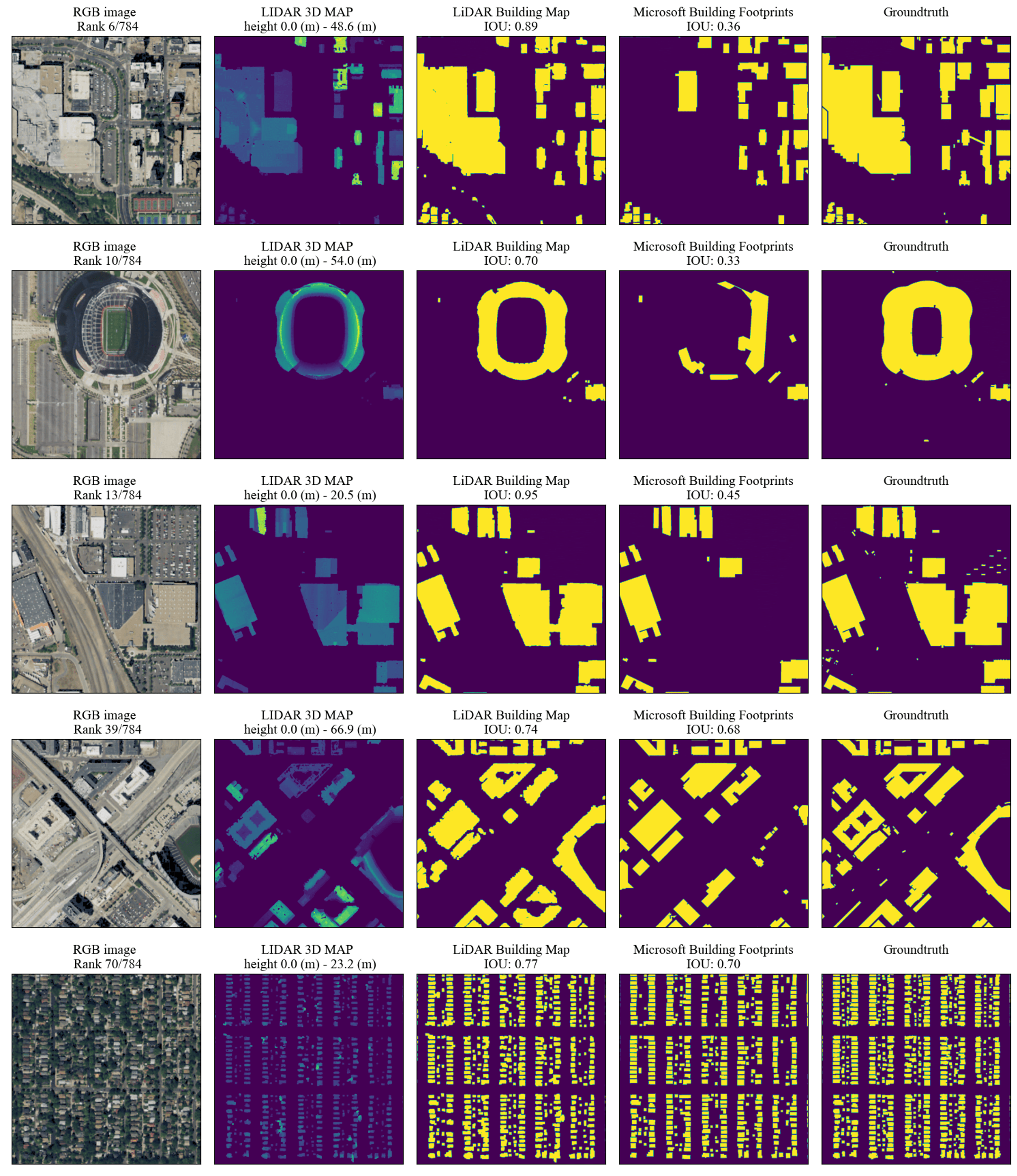}
 	\captionsetup{justification=centering}
	\caption{Five selected tiles and their building mapping results from the Denver dataset}
	\label{fig:fig5}
\end{figure*}

Among 784 tiles of Denver, we manually excerpted 5 tiles that rank high (top 10\%) and show notable differences in building maps. Figure \ref{fig:fig5} shows RGB images and their corresponding 3D building map, LiDAR building map, Microsoft Building Footprints, and ground-truth. The RGB imagery is from NAIP's orthoimagery. The vintage of the RGB imagery is 2015. Thus, we can expect the same buildings existing in both RGB image and LiDAR building map should also exist in Microsoft Building Footprints as the vintage of Microsoft Building Footprints (2018-2019) lies between those of RGB image (2015) and LiDAR data (2020). The ranking denoted with RGB image indicates the ranking of the largest difference between LiDAR building map and Microsoft building footprints. The ranking is to give a sense of how significant the difference between those building maps is in the entire 784 tiles of the Denver dataset. 3D building map generated by our workflow are also displayed for reference with its height range. IoU values calculated by comparing LiDAR building maps and Microsoft Building Footprints respectively to ground-truth are also provided.%Detailed description on the quality of 3D building map is provided in Section IV-C. IoU values calculated by comparing LiDAR building maps and Microsoft Building Footprints respectively to ground-truth are also provided.

Buildings that led the significant difference in their performances include huge buildings and unique-shape buildings. Microsoft Building Footprints were suffering particularly for large buildings like shopping malls or warehouses as shown in the first and third rows of Figure \ref{fig:fig5}. Also, buildings having unique shapes like sports complexes were not mapped properly as shown in the second and fourth rows of Figure \ref{fig:fig5}. These failures in Microsoft’s method can be regarded as a general limitation of deep learning-based supervised methods. Since large buildings and unique-shape buildings are not common, training samples for these kinds of buildings might not have been sufficient in the training data used to train the Microsoft's deep model. Another possible reason for the failure of Microsoft’s method could be the limited input size of image for the deep learning model. The typical input size of deep learning models for semantic segmentation is 256 by 256 pixels or 512 by 512 pixels. It means the model’s decision should be made by considering the limited area of 128-meter by 128-meter or 256-meter by 256-meter (assuming the GSD was 0.5-meter). For example, there is a chance that the input to a deep model contains only the middle of the roof of a large building, which cannot provide the model with enough information for a successful decision. Although the exact reason for the failure cannot be identified, an error in a large building is common in deep learning-based methods \cite{ji2018fully}.% On the other hand, our workflow produced reliable, accurate building maps regardless of building size and shape.

Our workflow obtained higher IoU than Microsoft's method not only for huge and unique-shape buildings but also for residential areas where small buildings dominate the tile as shown in the fifth row of Figure \ref{fig:fig5}. Our workflow extracts auxiliary units of houses such as garden sheds or detached garages well while many auxiliary units were not detected in Microsoft building footprints. This failure might be due to the limited resolution of the optical image used for generating building footprints or due to dense trees over residential buildings. Also, it could be due to the lack of training data for the auxiliary units. However, our workflow was able to detect auxiliary units that are generally not detected by Microsoft's method \cite{heris2020rasterized}. More discussion about the auxiliary unit is provided in Section III-C and Section IV-B.

\subsection{Results of the New York City dataset} \label{s3.2.}

Similar to Section III-A, LiDAR building map, Microsoft Building Footprints, and ground-truth were compared. The New York City dataset also has time discrepancies. The ground-truth of New York City is updated on a weekly basis by the DoITT team \cite{City_of_NewYork}. The vintage for this experiment is February 2022. For Microsoft Building Footprints, less than 5\% of total building labels had a time tag but most of them were 2019. LiDAR building map was generated with LiDAR data for 2014. The default parameter values were used for the entire area to produce the LiDAR building map.\newline

\subsubsection{Conventional quantitative results} \label{s3.2.1.}

LiDAR building map and Microsoft Building Footprints of the entire study area (357 {km\textsuperscript{2}}) of the New York City dataset were evaluated in terms of IoU, precision, recall, and F1-score (Table~\ref{tab:table2}). LiDAR building map obtained higher accuracy than Microsoft Building Footprints in all metrics except for precision. The low precision is mainly related to the building boundary refinement with dilation kernel K3. A more detailed explanation is given in Section IV-B. The result shows that building maps generated by our workflow are more accurate than Microsoft Building Footprints overall, even with a larger time gap to the ground-truth. Considering that our workflow used only a single default parameter set for the entire area mapping, the results show that our workflow is well-generalizable.
\begin{table}[ht]
	\caption{Conventional quantitative results of the New York City dataset}
	\centering
	\begin{tabular}{ccccc}
%	\begin{tabular}{lllll}	
		\toprule
% 		\multicolumn{2}{c}{Part}                   \\
% 		\cmidrule(r){1-2}
		& IoU     & Precision & Recall & F1-score \\
		\midrule
		Our workflow & \textbf{75.9} &  80.3 & \textbf{93.2} & \textbf{86.3}\\
		\midrule
		Microsoft's & 72.8 &  \textbf{84.6} & 83.9 & 84.3\\
%		LiDAR building map & 81.8 & 91.2 & 88.8 & 90.0\\
		\bottomrule
		
	\end{tabular}
	\label{tab:table2}
\end{table}
% \begin{table}[ht]
% 	\caption{Conventional quantitative results of the New York City dataset}
% 	\centering
% 	\begin{tabular}{ccccc}
% %	\begin{tabular}{lllll}	
% 		\toprule
% % 		\multicolumn{2}{c}{Part}                   \\
% % 		\cmidrule(r){1-2}
% 		& IoU     & Precision & Recall & F1-score \\
% 		\midrule
% 		LiDAR building map & \textbf{75.9} &  80.3 & \textbf{93.2} & \textbf{86.3}\\
% 		\midrule
% 		Microsoft Building Footprints & 72.8 &  \textbf{84.6} & 83.9 & 84.3\\
% %		LiDAR building map & 81.8 & 91.2 & 88.8 & 90.0\\
% 		\bottomrule
		
% 	\end{tabular}
% 	\label{tab:table2}
% \end{table}

\subsubsection{Tiling comparison} \label{s3.2.2.}
\begin{figure*}
	\centering
	\includegraphics[width=5.5in]{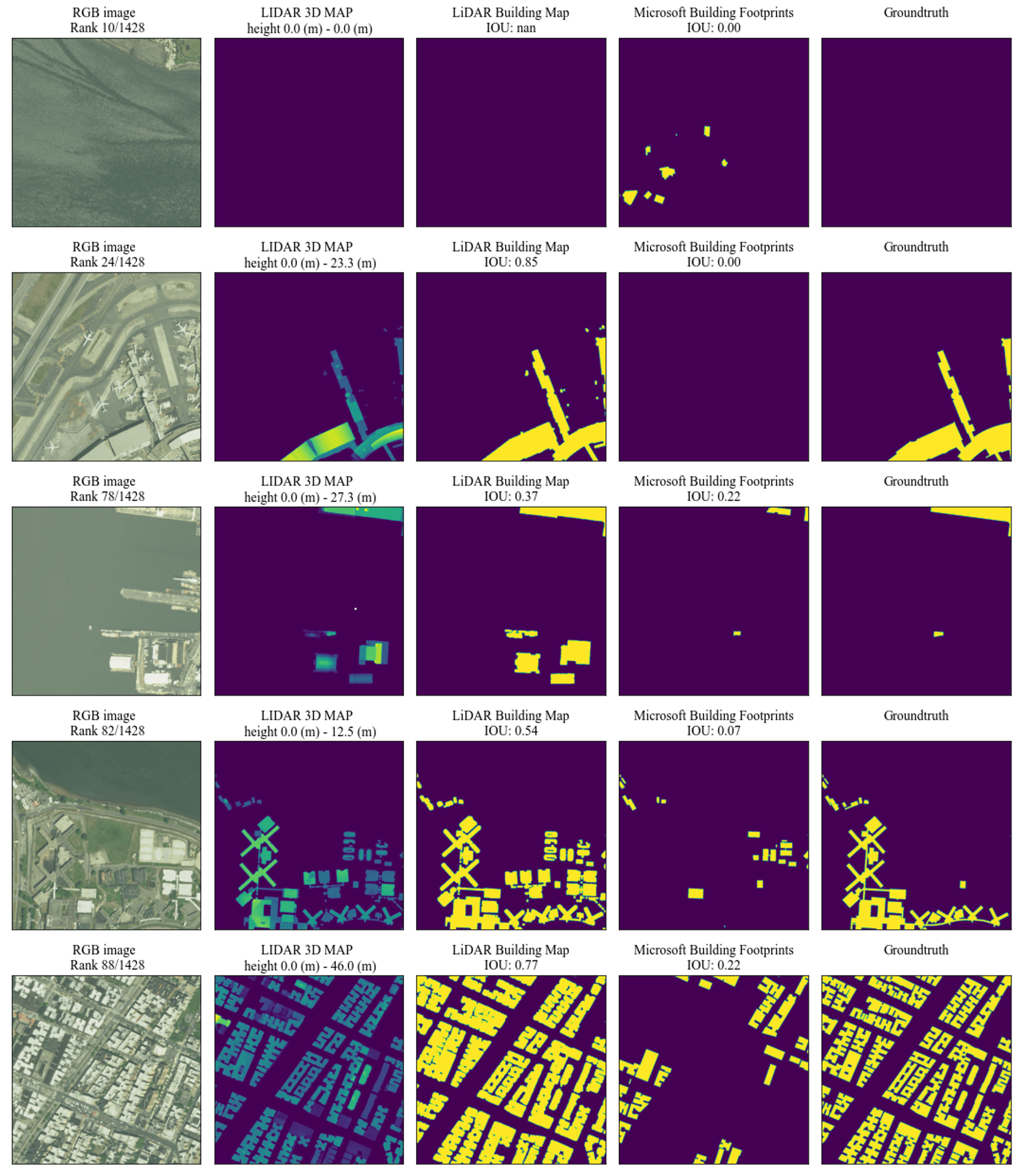}
 	\captionsetup{justification=centering}
	\caption{Five selected tiles and their building mapping results in the New York City dataset}
	\label{fig:fig6}
\end{figure*}

Among 1428 tiles of New York City, we displayed 5 tiles that rank high (top 10\%) and show notable differences in building maps. Figure \ref{fig:fig6} shows RGB images and their corresponding 3D building map, LiDAR building map, Microsoft Building Footprints, and ground-truth. 
The RGB imagery is from NAIP's orthoimagery. The RGB image was taken in 2015. Thus, we can expect the same buildings existing in both RGB image and ground-truth should exist in Microsoft Building Footprints as the vintage of Microsoft Building Footprints (2019) are between those of RGB image (2015) and ground-truth (2022). The ranking denoted with RGB image indicates the ranking of the largest difference between LiDAR building map and Microsoft building footprints. 3D building maps generated by our workflow are also displayed for reference with their height range. IoU values calculated by comparing LiDAR building maps and Microsoft Building Footprints respectively to ground-truth are also provided.

Unlike the Denver dataset, most of the errors in the top ranks were from the sea. Although the net areas of errors were not relatively large, Microsoft Building Footprints often had some artifacts in water bodies as shown in the first row of Figure \ref{fig:fig6}. Microsoft Building Footprints' errors over the sea were also noted by \cite{heris2020rasterized}. Microsoft's method might have had trouble differentiating between buildings and other objects such as clouds, shadows, or ships over the sea. LiDAR building map generated by our workflow rarely had errors in water bodies as the workflow includes the water body masking process. However, our workflow detected some large ships as buildings as shown in the third row of Figure \ref{fig:fig6}. Although not displayed in Figure \ref{fig:fig6}, the workflow sometimes produces artifacts near coastlines due to the DTM error. More discussion of DTM-related errors is provided in Section IV-C.

On land, similar to the results of the Denver dataset, most of the errors in Microsoft Building Footprints were large buildings and uniquely shaped buildings. One example is an airport as shown in the second row of Figure \ref{fig:fig6}. This is probably due to the lack of samples for airports in the training dataset of Microsoft's method. One unique error in the LiDAR building map is that it sometimes considers airplanes as buildings. This is because the physical properties of airplanes from airborne LiDAR data are very similar to buildings. Although not excerpted here, large trailers and cargos were other similar examples of misclassification in the LiDAR building map.% Section IV-D summarizes the potential error sources of building maps generated from our workflow.

Another common error source was the time discrepancies among datasets. In the fourth row of Figure \ref{fig:fig6}, the commission errors of the LiDAR building map are due to the time gap. The tile is from Rikers Island, New York. We found that some detected buildings (i.e., George Motchan Detention Center) in the LiDAR building map were demolished after the LiDAR scanning. Microsoft's method, however, failed to detect those buildings, which must have existed at the time of the mapping. This failure might be either due to the unique shape of the buildings or the non-availability of cloud-free optical imagery. 

The fifth row of Figure \ref{fig:fig6} shows a significant difference between the building maps in the downtown area. The error in Microsoft's case seems neither due to the time gap nor the shape of the buildings. It might be due to the defect in an optical image used for the mapping. Compared to this unexpected, inexplicable error from the image-based Microsoft's method, errors in LiDAR building maps are generally easy to be explained, and our workflow rarely has an error caused by the data itself. This is because our workflow focused on the physical property of the building instead of complex modeling that is not easily generalized. Also, as LiDAR uses an active sensor, the performance is rarely affected by clouds or other atmospheric conditions. It is a significant advantage of LiDAR-based building mapping over optical image-based building mapping. As errors from our workflow are generally more explainable and predictable than image-based building mappings, uncertainties of the building map could be significantly relieved and errors could be more controllable, which would be beneficial to subsequent analyses that use building maps. %Furthermore, even though acquiring optical images could be cheaper than performing ALS, having cloud-free high-resolution images over a large area would be very expensive and may be even impossible for timely applications, such as disaster management and humanitarian aid. 
One limitation of our workflow is that it can produce ``fat'' buildings, as shown in the fifth row of Figure \ref{fig:fig6}. This is related to the building boundary refinement process with dilation (K3), and this is the main reason why the precision of the LiDAR building map is lower than that of Microsoft Building Footprints in Table~\ref{tab:table2}. More discussion about K3 and its impact is provided in Section IV-B.

\subsection{Error analysis}\label{Error analysis} \label{s3.3.}

\subsubsection{Omission error} \label{s3.3.1.}

\begin{figure}
	\centering
	\includegraphics[width=2.75in]{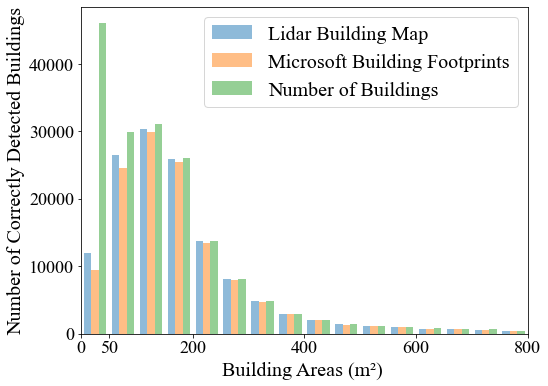}
 	%\captionsetup{justification=centering}
 	\caption{The numbers of correctly detected buildings and the number of buildings in the ground-truth according to the building area (0-800 {m\textsuperscript{2}}) in the Denver dataset}
	\label{fig:fig7}
\end{figure}

\begin{figure}
	\centering
	\includegraphics[width=2.75in]{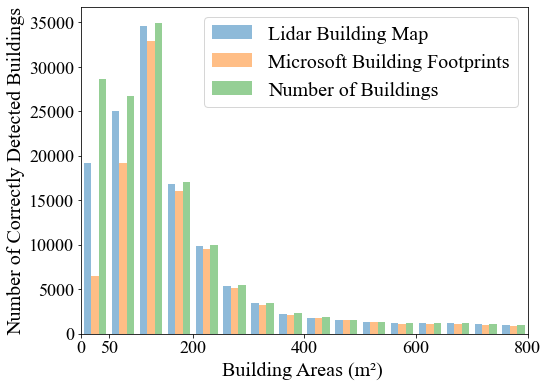}
 	%\captionsetup{justification=centering}
 	\caption{The numbers of correctly detected buildings and the number of buildings in the ground-truth according to the building area (0-800 {m\textsuperscript{2}}) in the New York City dataset}
	\label{fig:fig8}
\end{figure}

We observed both maps from our workflow and Microsoft’s method often miss small buildings. To evaluate the performance, focusing on small-to-medium-sized buildings (< 800 {m\textsuperscript{2}}), we counted the number of correctly detected buildings for different building areas. Here, the building area refers to the area of each building's footprint. We defined a ``correctly detected building'' as an instance that exists in the ground-truth and its overlapped portion with generated building instances is more than 50\%. Figure \ref{fig:fig7} and Figure \ref{fig:fig8} show the number of correctly detected buildings according to the building area for the Denver dataset and New York City dataset, respectively. The ``Number of Buildings'' in those graphs refers to the number of buildings in ground-truth for each building area category. %It shows the distribution of the number of buildings in the ground-truth according to the building area and the performance trends. %Quantitative results by building area are detailed in later paragraphs. 
Only buildings smaller than 800 {m\textsuperscript{2}} were illustrated as the performance significantly varied in relatively small buildings (< 100 {m\textsuperscript{2}}) and the trend was generally maintained for larger buildings.

For the Denver dataset, both maps miss many buildings smaller than 50 {m\textsuperscript{2}}, which are mostly accessory units such as storage sheds or detached garages. In the New York City dataset, however, the ratio of correctly detected buildings was significantly higher than that of the Denver dataset. This is because the authoritative map of New York City did not label storage sheds as a building class unlike that of Denver. Since our workflow uses an erosion with K1 $=7$, some buildings smaller than around 10 {m\textsuperscript{2}} (most of the storage sheds) were removed. Smaller buildings can be detected better with smaller K1, but the chance of detecting non-building small objects as a building class increases. The trade-off by K1 is detailed in Section IV-B.

The significantly different statistics between the two datasets indicate that authoritative maps are not consistent as building definitions may vary for different states and countries. Considering this issue, our workflow has a good potential to generate more consistent building map as it is based on definitive rules.

\begin{figure}
	\centering
	\includegraphics[width=3.5in]{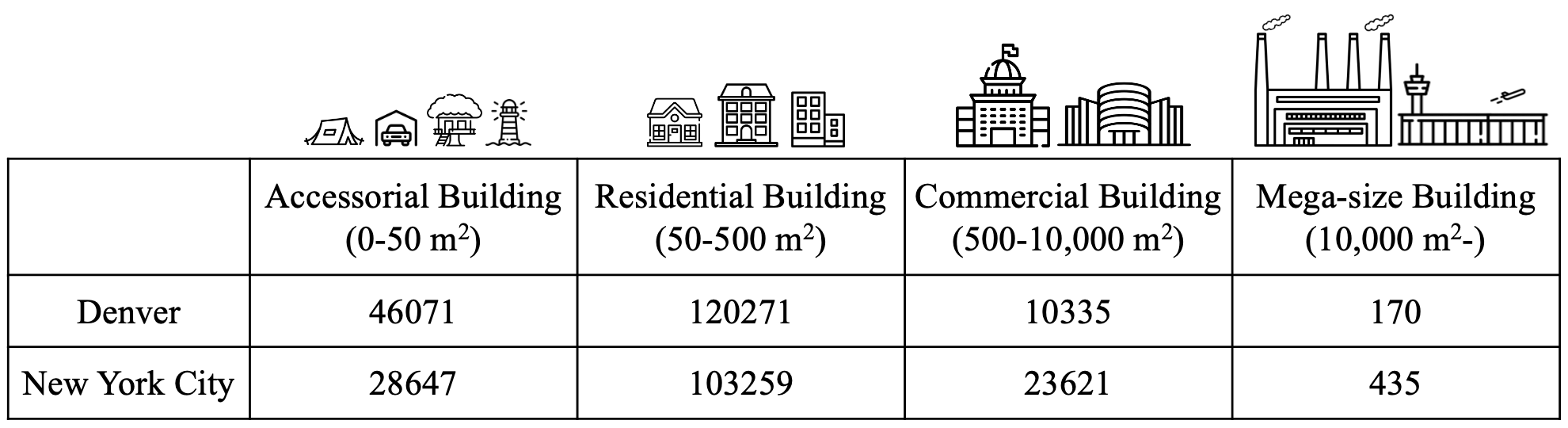}
%  	\captionsetup{justification=centering}
 	\caption{The number of buildings in ground-truths according to the building area (Icon copyright: Flaticon)}
	\label{fig:fig9}
\end{figure}

Figure \ref{fig:fig9} provides the building counts in ground-truths of the two datasets. Each category is to represent the different types of buildings. The categorization consists of four classes with different ranges of building area: 0-50 {m\textsuperscript{2}} (``accessorial class''), 50-500 {m\textsuperscript{2}} (``residential class''), 500-10,000 {m\textsuperscript{2}} (``commercial class''), and 10,000 {m\textsuperscript{2}}– (``mega-size class'').

The first category represents accessorial buildings. It can include storage sheds, detached garages, trailers, portable cabins, and so forth. Some building types in this category are often excluded in some classification systems.  Indeed, the authoritative map of Denver categorized the storage shed as a building while that of New York City excluded it. The second category, the residential class, represents most of the typical residential buildings. This category accounts for the majority of buildings. The commercial class, represents large commercial buildings. This category may include office buildings, retails, hospitals, warehouses, and industrial buildings. The last category represents mega-size buildings such as large shopping malls, factories, train stations, airports, and sports complexes. Based on this categorization, we compared the detection performances of our workflow and Microsoft’s method. Although the type of building cannot be classified by only their areas, this categorization can provide a sense of the difference in performance according to the building types.

Table~\ref{tab:table3} and Table~\ref{tab:table4} show the detection (true positive) rate of each category respectively for each dataset. The detection rate refers to the ratio of the number of correctly detected buildings to the number of buildings in ground-truth. 

\begin{table}%[b]
%  	\captionsetup{justification=centering}
 	\caption{Detection (true positive) rates according to the building category in the Denver dataset}
	\centering
	\begin{tabular}{ccccc}
%	\begin{tabular}{lllll}	
		\toprule
         &  \multicolumn{4}{c}{Detection rates (\%)}\\

        \midrule
% 		\multicolumn{2}{c}{Part}                   \\
% 		\cmidrule(r){1-2}
		& Accessorial-     & Residential- & Commercial- & Mega-size \\
		\midrule
		Our workflow & \textbf{25.8} &  \textbf{96.1} & \textbf{98.4} & \textbf{97.6}\\
		\midrule
		Microsoft's & 20.6 &  93.3 & 97.1 & 92.4\\
%		LiDAR building map & 81.8 & 91.2 & 88.8 & 90.0\\
		\bottomrule
		
	\end{tabular}
	\label{tab:table3}
\end{table}

% \begin{table*}%[b]
%  	\captionsetup{justification=centering}
%  	\caption{Detection (true positive) rates according to the building area in the Denver dataset}
% 	\centering
% 	\begin{tabular}{ccccc}
% %	\begin{tabular}{lllll}	
% 		\toprule
% % 		\multicolumn{2}{c}{Part}                   \\
% % 		\cmidrule(r){1-2}
% 		& 0-50 ({m\textsuperscript{2}})     & 50-500 ({m\textsuperscript{2}}) & 500-10,000 ({m\textsuperscript{2}}) & 10,000- ({m\textsuperscript{2}}) \\
% 		\midrule
% 		LiDAR building map & \textbf{25.8} &  \textbf{96.1} & \textbf{98.4} & \textbf{97.6}\\
% 		\midrule
% 		Microsoft Building Footprints & 20.6 &  93.3 & 97.1 & 92.4\\
% %		LiDAR building map & 81.8 & 91.2 & 88.8 & 90.0\\
% 		\bottomrule
		
% 	\end{tabular}
% 	\label{tab:table3}
% \end{table*}

\begin{table}%[b]
%  	\captionsetup{justification=centering}
 	\caption{Detection (true positive) rates according to the building area in the New York City dataset}
	\centering
	\begin{tabular}{ccccc}
%	\begin{tabular}{lllll}	
		\toprule
         &  \multicolumn{4}{c}{Detection rates (\%)}\\

        \midrule
% 		\multicolumn{2}{c}{Part}                   \\
% 		\cmidrule(r){1-2}
		& Accessorial-     & Residential- & Commercial- & Mega-size \\
		\midrule
		Our workflow & \textbf{67.1} &  \textbf{97.4} & \textbf{97.7} & \textbf{96.1}\\
		\midrule
		Microsoft's & 22.7 &  88.4 & 94.2 & 89.0\\
%		LiDAR building map & 81.8 & 91.2 & 88.8 & 90.0\\
		\bottomrule
		
	\end{tabular}
	\label{tab:table4}
\end{table}

% \begin{table*}%[b]
%  	\captionsetup{justification=centering}
%  	\caption{Detection (true positive) rates according to the building area in the New York City dataset}
% 	\centering
% 	\begin{tabular}{ccccc}
% %	\begin{tabular}{lllll}	
% 		\toprule
% % 		\multicolumn{2}{c}{Part}                   \\
% % 		\cmidrule(r){1-2}
% 		& 0-50 ({m\textsuperscript{2}})     & 50-500 ({m\textsuperscript{2}}) & 500-10,000 ({m\textsuperscript{2}}) & 10,000- ({m\textsuperscript{2}}) \\
% 		\midrule
% 		LiDAR building map & \textbf{67.1} &  \textbf{97.4} & \textbf{97.7} & \textbf{96.1}\\
% 		\midrule
% 		Microsoft Building Footprints & 22.7 &  88.4 & 94.2 & 89.0\\
% %		LiDAR building map & 81.8 & 91.2 & 88.8 & 90.0\\
% 		\bottomrule
		
% 	\end{tabular}
% 	\label{tab:table4}
% \end{table*}

In the Denver dataset, both methods performed poorly in the accessorial class. However, in the New York City dataset, the performance gap between ours and Microsoft's method was significant. Our method produced a 67.1\% detection rate while Microsoft’s method produced a 22.7\% detection rate in the accessorial class. The significant difference between two datasets is due to the difference in their classification system. As mentioned, storage sheds, which account for the majority of the accessorial class in Denver, were classified as building in Denver but not in New York City. Since most of the storage sheds were not detected well in both methods, the detection rates for accessorial buildings were poor in both methods. On the other hand, in the New York City dataset, as storage sheds (which are difficult to be detected by both methods) were not included in the building class, the performance gap between the two methods becomes evident.

Although the statistics of actual building type was unknown, we found a noteworthy difference between the distributions of the two datasets. 52.0\% of buildings from the accessorial class were less than 20 {m\textsuperscript{2}} in the Denver dataset, while only 17.9\% of accessorial buildings were less than 20 {m\textsuperscript{2}} in the New York City dataset. Considering that most of the storage sheds are smaller than 20 {m\textsuperscript{2}}, we can conclude the difference in their classification systems made the big performance gap in accessorial class.

Except for the accessorial category, our workflow produced a detection rate of over 95.0\% in all cases. Microsoft’s method also produced a good performance in general, but our workflow outperformed Microsoft's method.\newline

\subsubsection{Commission error} \label{s3.3.2.}

Similar to the omission error, we tabulated commission (false positive error) rates of the two methods per building class. Table~\ref{tab:table5} and Table~\ref{tab:table6} show the commission rates respectively for the two datasets. The commission rate refers to the number of incorrectly detected buildings out of the total number of buildings in ground-truth. The ``incorrectly detected building'' is defined as an instance that exists in generated building map but its overlapped portion with the ground-truth is less than 50\%. 

As a result, our workflow obtained higher commission errors than Microsoft's method. Particularly, the commission rate for the accessorial class in the New York City dataset is larger than others. This is closely associated with the fact that the LiDAR building map showed a high detection rate in the accessorial class. Simply put, as our workflow detected accessorial buildings much more than Microsoft’s method, the LiDAR building map has more positives (either true or false) than Microsoft Building Footprints in the accessorial class.

Commission error accounts for a relatively small portion of the total error in both methods. The time discrepancy to ground-truth accounts for some portions of the reason. Except for this, most of the commission errors from our workflow were overhanging trees or DTM-related artifacts. Detailed analyses of these errors are described in Section IV-C. 

\begin{table}[ht]
%  	\captionsetup{justification=centering}
 	\caption{Commission (false positive) rates according to the building area in the Denver dataset}
	\centering
	\begin{tabular}{ccccc}
%	\begin{tabular}{lllll}	
		\toprule
         &  \multicolumn{4}{c}{Commission rates (\%)}\\

        \midrule
% 		\multicolumn{2}{c}{Part}                   \\
% 		\cmidrule(r){1-2}
		& Accessorial-     & Residential- & Commercial- & Mega-size \\
		\midrule
		Our workflow & 4.3 &  2.6 & 2.8 & 1.2\\
		\midrule
		Microsoft's & \textbf{2.4} &  \textbf{0.6} & \textbf{0.8} & \textbf{0.6}\\
		\bottomrule
		
	\end{tabular}
	\label{tab:table5}
    \vspace{-0.10in}
\end{table}

\begin{table}[ht]
%  	\captionsetup{justification=centering}
 	\caption{Commission (false positive) rates according to the building area in the New York City dataset}
	\centering
	\begin{tabular}{ccccc}
%	\begin{tabular}{lllll}	
		\toprule
         &  \multicolumn{4}{c}{Commission rates (\%)}\\

        \midrule
% 		\multicolumn{2}{c}{Part}                   \\
% 		\cmidrule(r){1-2}
		& Accessorial-     & Residential- & Commercial- & Mega-size \\
		\midrule
		Our workflow & 19.1 &  4.6 & 2.5 & 6.0\\
		\midrule
		Microsoft's & \textbf{6.5} &  \textbf{1.5} & \textbf{1.3} & \textbf{1.8}\\
		\bottomrule
		
	\end{tabular}
	\label{tab:table6}
    \vspace{-0.10in}
\end{table}
%\label{sec:others}

% The poor performance in the accessorial category is partly due to the definition of the correctly detected building in our study. We defined the correctly detected building as an instance that exists in ground-truth and its overlapped portion with generated building instances is more than 50\%. Generally, since errors in building footprints tend to occur near the building boundary, the definition of the correctly detected building leads to a harsher standard for small buildings as the proportion of the building boundary is necessarily larger for small buildings than for large buildings.

\subsection{Summary of results} \label{s3.4.}
Quantitative and qualitative analyses with extensive experimental areas showed that our workflow produced more accurate results than Microsoft Building Footprints across most criteria, despite a larger time gap with the ground-truth vintage. To be specific, our workflow produced higher IoU, recall, and F1-score in both datasets. Our workflow’s detection rates were higher than that of Microsoft’s method in all categories of building. Also, we found that our workflow was generally better at detecting large and uniquely shaped buildings than Microsoft's method. However, our workflow produced lower precision in the New York City dataset and had higher commission rates in all building categories than Microsoft's method.

Notably, our workflow is an unsupervised method and the results we obtained were with a single set of default parameters for an entire 550 {km\textsuperscript{2}} area. Despite its simplicity and ease of use, our workflow generally produced superior results to Microsoft's method, which relied on millions of training labels. Moreover, while errors in our workflow can be readily explained, those in Microsoft's method are not as straightforward.

These results suggest that, with quality airborne LiDAR data, our workflow can serve as a viable alternative to deep learning-based methods using optical images for 2D and 3D building mapping. Furthermore, our comprehensive analyses of large-scale building maps from different methods illuminate the strengths and weaknesses of each approach, providing valuable insights into what can be expected from major large-area building mapping techniques.

% \begin{table*}%[ht]
%  	\captionsetup{justification=centering}
%  	\caption{Commission (false positive) rates according to the building area in the Denver dataset}
% 	\centering
% 	\begin{tabular}{ccccc}
% %	\begin{tabular}{lllll}	
% 		\toprule
% % 		\multicolumn{2}{c}{Part}                   \\
% % 		\cmidrule(r){1-2}
% 		& 0-50 ({m\textsuperscript{2}})     & 50-500 ({m\textsuperscript{2}}) & 500-10,000 ({m\textsuperscript{2}}) & 10,000- ({m\textsuperscript{2}}) \\
% 		\midrule
% 		LiDAR building map & 4.3 &  2.6 & 2.8 & 1.2\\
% 		\midrule
% 		Microsoft Building Footprints & \textbf{2.4} &  \textbf{0.6} & \textbf{0.8} & \textbf{0.6}\\
% 		\bottomrule
		
% 	\end{tabular}
% 	\label{tab:table5}
% \end{table*}

% \begin{table*}%[ht]
%  	\captionsetup{justification=centering}
%  	\caption{Commission (false positive) rates according to the building area in the New York City dataset}
% 	\centering
% 	\begin{tabular}{ccccc}
% %	\begin{tabular}{lllll}	
% 		\toprule
% % 		\multicolumn{2}{c}{Part}                   \\
% % 		\cmidrule(r){1-2}
% 		& 0-50 ({m\textsuperscript{2}})     & 50-500 ({m\textsuperscript{2}}) & 500-10,000 ({m\textsuperscript{2}}) & 10,000- ({m\textsuperscript{2}}) \\
% 		\midrule
% 		LiDAR building map & 19.1 &  4.6 & 2.5 & 6.0\\
% 		\midrule
% 		Microsoft Building Footprints & \textbf{6.5} &  \textbf{1.5} & \textbf{1.3} & \textbf{1.8}\\
% 		\bottomrule
		
% 	\end{tabular}
% 	\label{tab:table6}
% \end{table*}
% %\label{sec:others}

\section{Discussion} \label{s4.}
\subsection{Overview} \label{s4.1.}
% The results in Section III demonstrated that our workflow can produce accurate 2D and 3D building maps from airborne LiDAR data compared to Microsoft Building Footprints. %The results in Section 3 demonstrated the great potential of our building mapping method that leverages LiDAR and an unsupervised method. 
% Despite its robust performance, we observed some limitations and errors in our building maps. With more exhaustive experiments, 
% We found some errors could be prevented with parameter tunings, but some limitations are difficult to avoid. 
This section elaborates on the impact of parameter selections and the limitations of our workflow. To explore these more in-depth, we added other study areas outside of Denver and New York City for more diversity. Added areas include other large cities, small towns in tropical regions, uninhabited natural areas, and so forth. In Section IV-B, we discussed some influential parameters and showed how they affect the results and how they can be adjusted to reduce errors. In Section IV-C, we detailed limitations that were not easily resolved even with parameter tunings. %Lastly, in Section IV-D, we discussed the scalability of our workflow for global-scale mapping.
% Lastly, in Section IV-D, we discussed the pros and cons of our workflow compared to image-based mapping.

\subsection{Suggestion for parameter tuning} \label{s4.2.}
We have found that proper parameter tuning can improve its performance even more in some cases. Examples include the case when one has a certain preference for the accessorial class (related to parameter K1) and the case when one tries to map an area with forested areas (related to parameter DT). Also, we found adjusting building boundaries (related to parameter K3) can significantly affect the quantitative metrics. Here, we discuss these parameters and provide general tips for parameter tuning.\newline

\subsubsection{K1 parameter tuning for handling accessorial buildings} \label{s4.2.1}
Generally, as K1 gets larger, commission error decreases while omission error increases. This is because both small buildings and non-building small objects are more likely to be removed during erosion with a larger K1 value. To be specific, small buildings whose width is shorter than K1$*$GSD will be removed by the erosion. As typical residential houses are larger than 3.5-meter by 3.5-meter, default K1 $=7$ for 0.5-m GSD will not remove usual residential buildings. However, some accessorial buildings like detached garages and garden sheds can be removed with the default K1 value. The erosion kernel removes most of the trees. This is because the laser penetrates the trees and observes the ground beneath the tree, and eventually makes trees like salt-and-pepper noise-like small patches. However, some dense trees or shrubs that do not allow penetration and are represented as a contiguous group of patches can still remain after the erosion. This is why the planarity-based filtering with DT is needed.%, and actually, it removes most trees left even after the morphological filtering.% However, there are some rare cases where dense trees still remain. Particularly if dense trees are connected to buildings, those trees are likely to remain after the morphological filtering and the planarity-based filtering as their planarity values are likely to be as high as buildings due to the contiguous building. Dense trees contiguous to buildings were one of the most common commission errors of our workflows.

\begin{figure}
	\centering
	\includegraphics[width=3.5in]{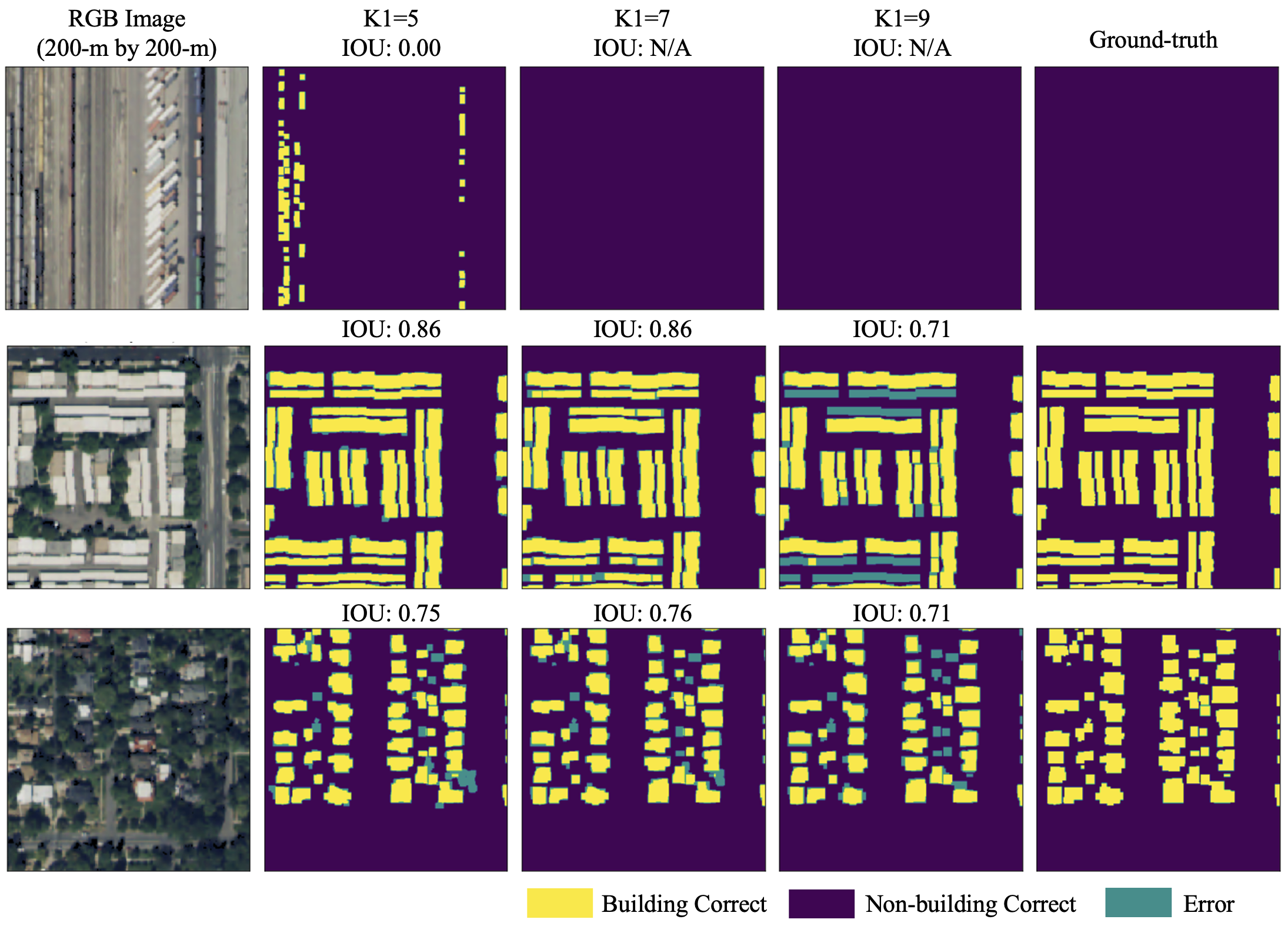}
	\caption{Generated building maps with different K1 values}
	\label{fig:fig10}
    \vspace{-0.15in}
\end{figure}

Figure \ref{fig:fig10} shows three sample areas from Denver that showed notable differences for different K1 values. ``Building correct'' refers to the pixel which both LiDAR building map and ground-truth classified as building. ``Non-building correct'' refers to the pixel which both classified as non-building. ``Errors'' indicates the pixel where LiDAR building map and ground-truth do not match. We found K1 $=7$ produces the highest performance in general, but there are pros and cons to using different K1 values. The first row of Figure \ref{fig:fig10} shows an example of a commission error that can occur when K1 is too small. Containers were not labeled as building in the Denver ground-truth, but K1 $=5$ detected some containers as buildings because some containers’ widths are larger than 2.5-meter. The second row of Figure \ref{fig:fig10} shows the example where K1 $=9$ was not able to detect detached garages as buildings while both smaller K1s detected them as buildings. The third row of Figure \ref{fig:fig10} show the example of storage sheds. The smaller the value of K1, the more often the storage shed was classified as a building. It shows the example of overhanging dense trees as well. With the smaller K1, overhanging dense trees are more likely to be detected as buildings.

% \begin{figure*}
% 	\centering
% 	\includegraphics[width=6in]{figures/fig10.png}
%  	\captionsetup{justification=centering}
% 	\caption{LiDAR building maps with different K1 values}
% 	\label{fig:fig10}
% \end{figure*}

We also quantitatively evaluated the impact of K1 by mapping the Denver dataset with different K1 values. Table~\ref{tab:table7} and Table~\ref{tab:table8} show the detection and commission rates, respectively. The accuracy significantly varied in accessorial class (0-50 {m\textsuperscript{2}}) for different K1 values. With the smaller K1, the detection rate increased but the commission rate increased as well. In other words, the smaller K1 is more likely to map small, accessorial buildings, but at the same time, it is more likely to classify non-building small objects as buildings. In contrast, the larger K1 could prevent the commission error but may miss some small buildings.
\begin{table}[ht]
%  	\captionsetup{justification=centering}
 	\caption{The impact of parameter K1 on the workflow's detection (true positive) rates in the Denver dataset}
	\centering
	\begin{tabular}{ccccc}
%	\begin{tabular}{lllll}	
		\toprule
         &  \multicolumn{4}{c}{Detection rates (\%)}\\

        \midrule
% 		\multicolumn{2}{c}{Part}                   \\
% 		\cmidrule(r){1-2}
		& Accessorial-     & Residential- & Commercial- & Mega-size \\
		\midrule
		K1=5 & \textbf{40.5} &  \textbf{97.4} & \textbf{98.5} & \textbf{97.6}\\
		\midrule
		K1=7(default) & 25.8 &  96.1 & 98.4 & \textbf{97.6}\\
		\midrule
		K1=9 & 13.2 &  91.5 & 98.1 & \textbf{97.6}\\
%		LiDAR building map & 81.8 & 91.2 & 88.8 & 90.0\\
		\bottomrule
		
	\end{tabular}
	\label{tab:table7}
\end{table}

\begin{table}[ht]
%  	\captionsetup{justification=centering}
 	\caption{The impact of parameter K1 on the workflow's commission (false positive) rates in the Denver dataset}
	\centering
	\begin{tabular}{ccccc}
%	\begin{tabular}{lllll}	
		\toprule
         &  \multicolumn{4}{c}{Commission rates (\%)}\\

        \midrule
% 		\multicolumn{2}{c}{Part}                   \\
% 		\cmidrule(r){1-2}
		& Accessorial-     & Residential- & Commercial- & Mega-size \\
		\midrule
		K1=5 & 12.9 &  3.5 & 3.2 & \textbf{1.2}\\
		\midrule
		K1=7(default) & 4.3 &  2.6 & 2.8 & \textbf{1.2}\\
		\midrule
		K1=9 & \textbf{1.9} &  \textbf{2.0} & \textbf{2.6} & \textbf{1.2}\\
%		LiDAR building map & 81.8 & 91.2 & 88.8 & 90.0\\
		\bottomrule
		
	\end{tabular}
	\label{tab:table8}
    \vspace{-0.15in}
\end{table}

Thus, our workflow can adjust the resulting building map to fit the classification scheme by tuning K1, unlike deep learning approaches, which may require relabeling the training data for different classification schemes. Moreover, the results suggest that the building map produced from our workflow could be a good alternative to the authoritative building map in terms of consistency and scalability. This is because while the criteria can be different in authoritative maps as we observed the difference between Denver and New York City, our workflow could provide more consistent results based on definitive rules, particularly if the building area is the main consideration of the classification system.\newline

\subsubsection{Dense trees (DT) parameter tuning for forested areas} \label{s4.2.2}

\begin{figure}
	\centering
	\includegraphics[width=3.5in]{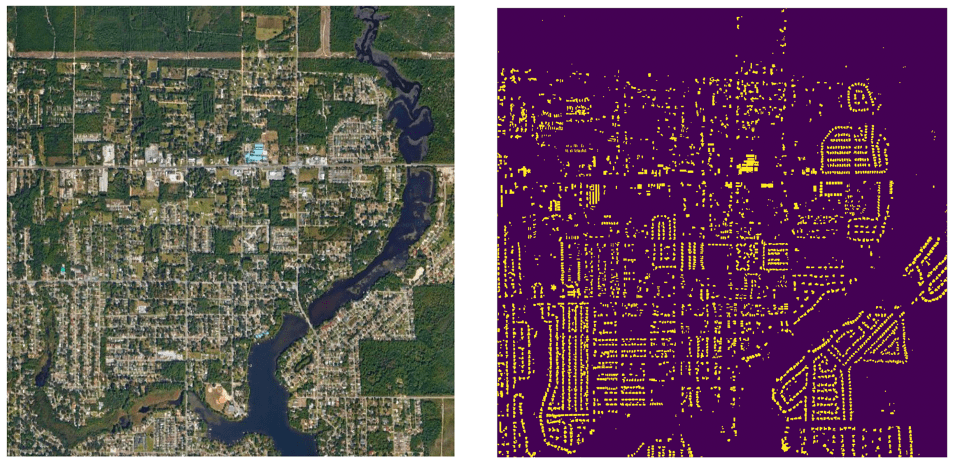}
	\caption{Satellite image and LiDAR building map of the Callaway dataset (3.3-km by 3.3-km)}
	\label{fig:fig11}
\end{figure}

The parameter DT is the threshold for the planarity-based filtering algorithm which is mainly for removing dense trees. As the value of DT increases, dense trees are likely to be removed, but buildings under overhanging trees are also likely to be removed. Here, we investigated this trade-off with the case of a town in Callaway, Florida, US. This area is around 11 {km\textsuperscript{2}} and includes residential areas and tropical evergreen trees with densely multi-layered leaves. The scanning was conducted with a Riegl Q-1560 system, and the point density was around 7-points/{m\textsuperscript{2}}. As the ALS was conducted from April to May of 2017, we can expect a fully leaf-on condition. Figure \ref{fig:fig11} shows the satellite image and building map generated by our workflow with the default parameter set.

\begin{figure}
	\centering
	\includegraphics[width=3in]{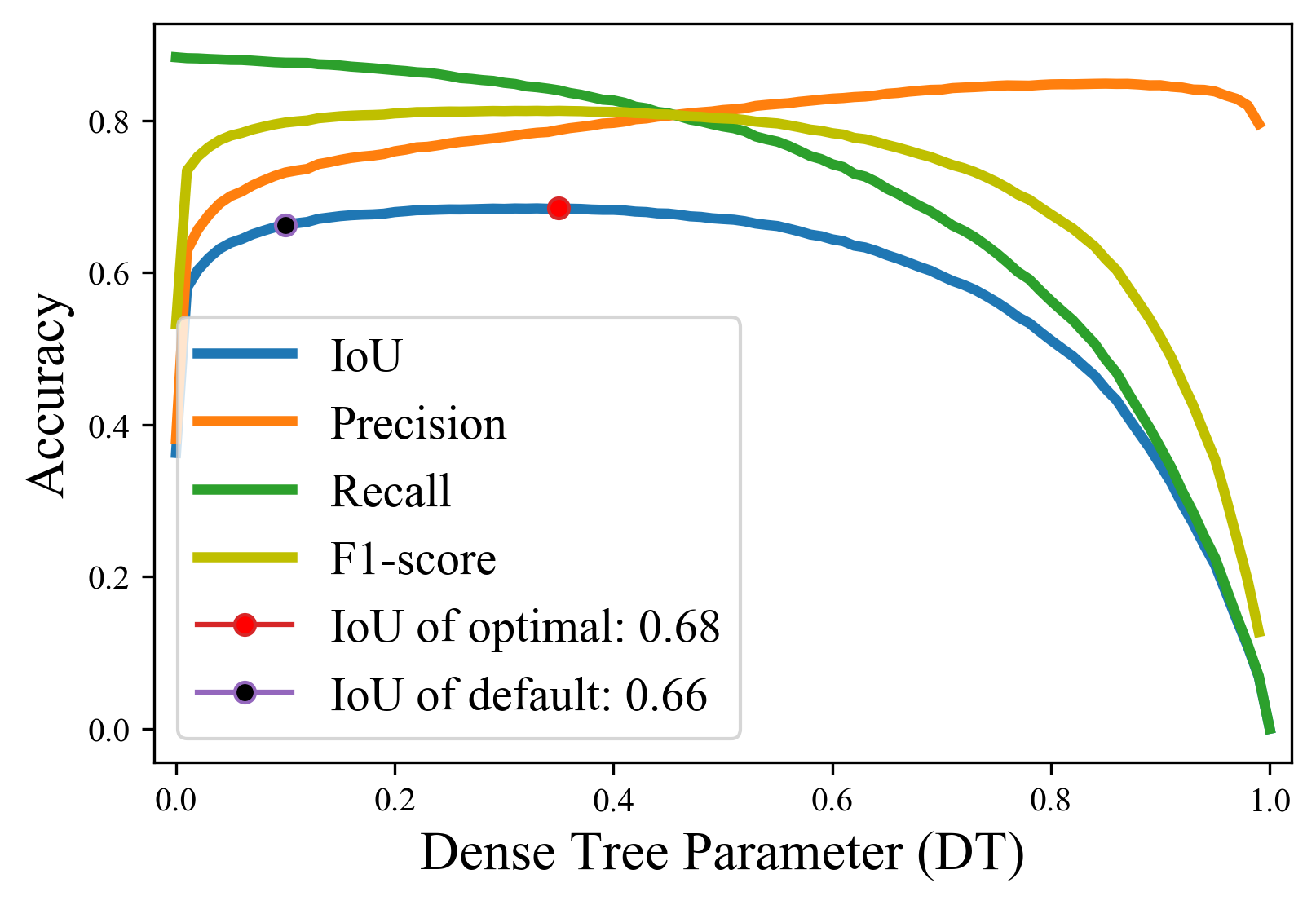}
	\caption{IoU, precision, recall, and F1-score according to the parameter DT in the Callaway dataset}
	\label{fig:fig12}
    \vspace{-0.15in}
\end{figure}

To explore the impact of DT values, we calculated quantitative metrics by changing DT values. IoU, precision, recall, and F1-score with different DT values are plotted in Figure \ref{fig:fig12}. For the calculation, we regarded Microsoft Building Footprints as the ground-truth as a more reliable ground-truth was not available over the Callaway area. The results turned out that the optimal value for DT in this study area was 0.35 and produced 0.68 IoU, while the default value of DT $=0.1$ produced 0.66 IoU. It indicates that proper selection of DT can increment accuracy slightly, but the default value still can produce reasonable accuracy. Also, the plateau in the graph of IoU between DT values of 0.05-0.5 was observed. It suggests the default DT value would produce near-optimal results in general cases and that the performance is not susceptible to DT as long as DT is in the proper range.

\begin{figure}
	\centering
	\includegraphics[width=3.5in]{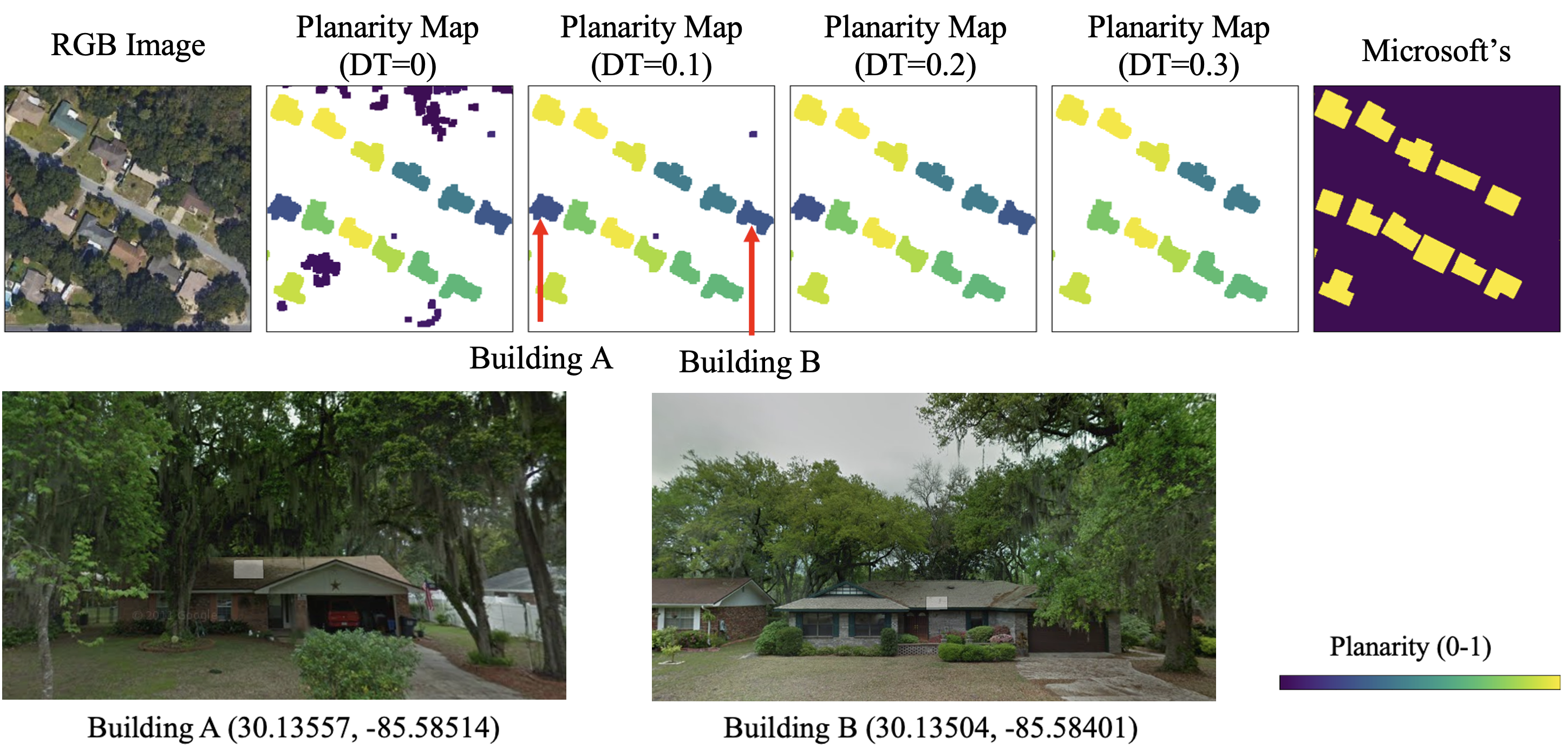}
 	%\captionsetup{justification=centering}
 	\caption{Generated building maps (planarity maps) according to the parameter DT and examples of buildings under overhanging trees in Google Street View (Image copyright: Google Inc.). The numbers in parentheses indicate the latitude and longitude of the building in decimal degrees, respectively.}
	\label{fig:fig13}
\end{figure}

Although DT $=0.35$ produced the highest IoU value, it does not necessarily guarantee the best building map. Figure \ref{fig:fig13} shows the subset of the study area in Callaway, Florida, US. The figure illustrates the generated maps according to different DT values. The planarity map represents the planarity value of each building candidate. The figure with DT $=0$ shows all building candidates before the planarity-based filtering. With the default value (DT $=0.1$), most dense trees were removed, but some non-building small objects remained. When DT $=0.2$, only building objects were extracted. However, when DT $=0.3$, two actual building objects with lower planarity values were removed. We found that two removed buildings were under overhanging trees through Google Street View. Thus, for this particular scene, the best DT was 0.2. It suggests that the best DT value can be different according to the study area. Moreover, especially in the tropical regions, we observed several cases where a building under overhanging trees has a lower planarity value than the dense trees. This reversal case happens more often when the point density is low. It remains as a limitation. An experiment on the impact of the point density is provided in Section IV-C.\newline

% \begin{figure*}
% 	\centering
% 	\includegraphics[width=5.5in]{figures/fig13.png}
%  	%\captionsetup{justification=centering}
%  	\caption{Examples of building maps (planarity maps) in the Callaway dataset according to the parameter DT and examples of buildings under overhanging trees in Google Street View (Image copyright: Google Inc.). The numbers in parentheses indicate the latitude and longitude of the building in decimal degrees, respectively.}
% 	\label{fig:fig13}
% \end{figure*}

%We also used our workflow on some uninhabited dense forests to investigate how many dense trees are classified as a building class. Study areas and LiDAR data collection dates are listed as follows: Monongahela National Forest (9 {km\textsuperscript{2}}, November – December of 2016), West Virginia, US, Yellowstone National Park (4.5 {km\textsuperscript{2}}, September–October of 2020), Wyoming, US, and Apalachicola National Forest (9 {km\textsuperscript{2}}, March-May of 2018), Florida, US. Since these areas are development restricted areas, we can assume there is no building. As a result of the default parameter, the Monongahela National Forest dataset results in 8 false building objects, the Yellowstone National Park dataset results in 31 false building objects, and the Apalachicola National Forest dataset results in 11 false building objects. The total area of false building objects was less than 0.5\% of entire areas in all cases. Although not all extracted objects were identified, most of them were trees or huge rocks. Also, 68\% of them were less than 50 {m\textsuperscript{2}}, which belongs to the category of the accessorial building class. The results indicate that our workflow even with the default parameter rarely produces false-positive errors in densely forested areas. However, if the experimental area includes a very large forested area, parameter tuning (e.g., using larger K1 and DT) is recommended to suit the purpose of the experiment.\newline

\subsubsection{K3 parameter tuning for refining building boundaries} \label{s4.2.3}
The K3 is for refining building boundary noise and recovering underestimated building area. LiDAR observations near building boundaries are mixed observations of ground points and building edges. This is because a typical airborne LiDAR scanner collects data points by emitting laser pulse radially with a whiskbroom pattern. Since the scan angle is not always perpendicular to the object on the ground, both the side and floor height values of the building can be collected as the height values of the building boundary.

\begin{figure}
    \vspace{-0.1in}
	\centering
	\includegraphics[width=2.75in]{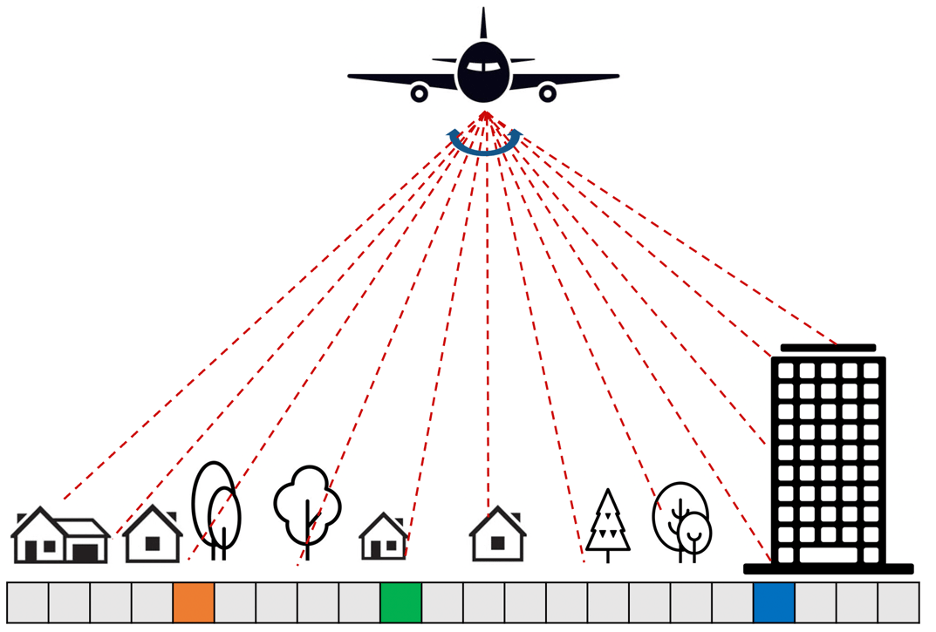}
	\caption{Graphical representation of airborne laser scanning}
	\label{fig:fig14}
    \vspace{-0.15in}
\end{figure}

Figure \ref{fig:fig14} shows a toy example of an ALS. The blue grid will have multiple points with mixed elevations of the bottom, side, and top of the building. Among these mixed elevations, we take the bottom elevation for its DSM value as described in Section II-B. Taking the lowest value, however, can result in the underestimation of the building areas. Even if multiple points are not acquired, the building area can be underestimated. For example, the green grid will be classified as a non-building pixel since the laser measures the bottom of the building. The case of the orange grid will have the same result, and it will cause the underestimation of the building boundary. Of course, if LiDAR collects the point from the edge of the roof, the area can be overestimated up to the one-pixel GSD. However, taking the lowest elevation underestimates the building area overall. In fact, the thickness of the underestimated boundary is affected by several variables, including the point density, the scan angle, and the relative position between the laser beam firing point and each object. Ideally, K3 must be tuned for each pixel considering those variables. However, some required variables are not provided in most laser scanning data, and calculating optimized K3 for every pixel will greatly increase the computation.

\begin{figure}
	\centering
	\includegraphics[width=3.5in]{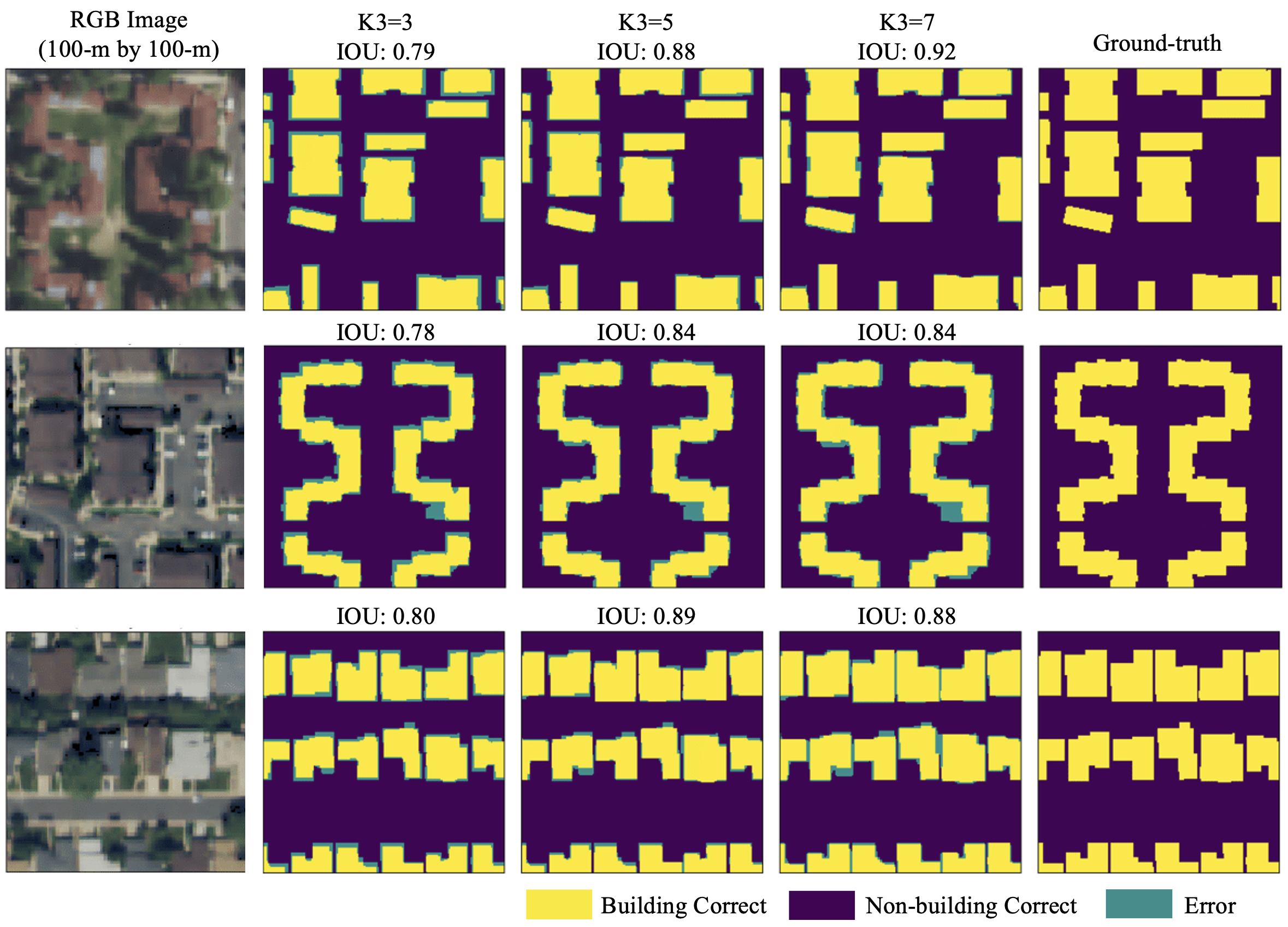}
%  	\captionsetup{justification=centering}
	\caption{Generated building maps with different K3 values}
	\label{fig:fig15}
    \vspace{-0.15in}
\end{figure}

Based on the quantitative evaluation, we confirmed the default K3 $=5$ produced high accuracy in general. However, we found that the K3 value significantly affects the accuracy, and the optimal value varies depending on study areas. Table~\ref{tab:table9} and Table~\ref{tab:table10} show the quantitative results for different K3 values from the Denver and New York City datasets, respectively. K3 $=5$ obtained the highest IoU in the Denver dataset while K3 $=3$ obtained the highest IoU in the New York City dataset. Figure \ref{fig:fig15} provides examples of boundary errors with different K3 values in the Denver dataset.

Inconsistent optimal K3 values can be attributed to several reasons. First, as described early, the optimal K3 should be determined by the function of the scan angle and the relative position between the sensor and the object. However, these conditions are not uniform due to the nature of the ALS data acquisition mechanism, which makes it challenging to optimize K3 value globally. Second, the ground-truth itself has an inconsistency. Since the ground-truth was labeled by multiple surveyors using images of different specifications, optical images might have not lied in the same condition in terms of their resolutions and ortho-rectification qualities. Also, there could have been some human biases and errors. Our workflow has a limitation in that it currently has a global K3 value and it cannot apply different sizes of K3 for individual objects.

\begin{table}[ht]
	\caption{The impact of parameter K3 on the workflow's quantitative metrics in the Denver dataset}
	\centering
	\begin{tabular}{ccccc}
%	\begin{tabular}{lllll}	
		\toprule
% 		\multicolumn{2}{c}{Part}                   \\
% 		\cmidrule(r){1-2}
		& IoU     & Precision & Recall & F1-score \\
		\midrule
		K3=1 & 70.1 &  \textbf{95.6} & 72.4 & 82.4\\
		\midrule
		K3=3 & 77.6 &  94.2 & 81.4 & 87.4\\
		\midrule
		K3=5(default) & \textbf{81.8} &  91.2 & 88.8 & \textbf{90.0}\\
		\midrule
		K3=7 & 80.5 &  85.6 & \textbf{93.2} & 89.2\\
%		LiDAR building map & 81.8 & 91.2 & 88.8 & 90.0\\
		\bottomrule
		
	\end{tabular}
	\label{tab:table9}
    \vspace{-0.1in}
\end{table}

\begin{table}[ht]
	\caption{The impact of parameter K3 on the workflow's quantitative metrics in the New York City dataset}
	\centering
	\begin{tabular}{ccccc}
%	\begin{tabular}{lllll}	
		\toprule
% 		\multicolumn{2}{c}{Part}                   \\
% 		\cmidrule(r){1-2}
		& IoU     & Precision & Recall & F1-score \\
		\midrule
		K3=1 & 77.0 &  \textbf{89.4} & 84.7 & 87.0\\
		\midrule
		K3=3 & \textbf{77.9} &  85.3 & 90.0 & \textbf{87.6}\\
		\midrule
		K3=5(default) & 75.9 &  80.3 & 93.2 & 86.3\\
		\midrule
		K3=7 & 72.2 &  75.1 & \textbf{94.8} & 83.9\\
%		LiDAR building map & 81.8 & 91.2 & 88.8 & 90.0\\
		\bottomrule
		
	\end{tabular}
	\label{tab:table10}
    \vspace{-0.15in}
\end{table}

% \begin{figure*}
% 	\centering
% 	\includegraphics[width=6in]{figures/fig15.png}
%  	\captionsetup{justification=centering}
% 	\caption{LiDAR building maps according to different K3 values}
% 	\label{fig:fig15}
% \end{figure*}

% \subsubsection{Other parameter tunings} \label{s4.2.4}

% Other parameters that can be tuned in our workflow include the spatial resolution of the building map, water buffer, and height threshold. We adopted the 0.5-meter resolution for all experiments in this paper because the ground-truth or Microsoft Building Footprints were made mostly based on a typical high-resolution satellite image, which has around 0.5-meter resolution. Any finer resolutions are always possible with the fine rasterization method as long as high point density LiDAR data and computational resources are available. Lastly, if there is any preferred height range for building classification, choosing different HT values instead of the default HT of 1.5-meter would be beneficial.

\subsection{Limitation} \label{s4.3}

Most artifacts from our workflow can be resolved to some extent by proper parameter tuning as discussed in Section IV-B. Also, most of the limitations are essentially associated with the trade-off between commission and omission errors. However, some limitations are not directly associated with the trade-off. This section devotes to discussing those limitations.\newline

\subsubsection{Error near skyscrapers} \label{s4.3.1}

\begin{figure}[ht]
    \vspace{-0.15in}
	\centering
	\includegraphics[width=3.5in]{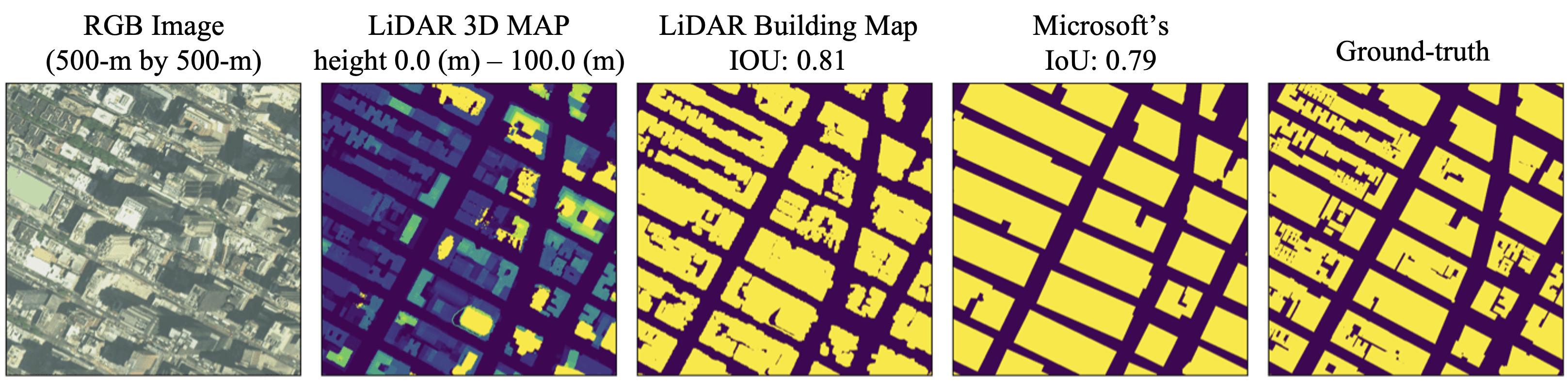}
%  	\captionsetup{justification=centering}
 	\caption{Errors due to occlusion caused by skyscrapers}
	\label{fig:fig16}
     \vspace{-0.1in}
\end{figure}

Unlike the optical image data, LiDAR is not affected by shadow. However, it still can be affected by occlusion as LiDAR emits laser pulse having a slant scan angle. This property can result in cases where LiDAR points were not collected in-between closely clustered tall buildings. This often occurs near skyscrapers. Figure \ref{fig:fig16} shows an example of Manhattan, New York. The occluded area due to skyscrapers resulted in either blurred boundaries or holes in the building map.\newline

% \begin{figure*}
% 	\centering
% 	\includegraphics[width=6in]{figures/fig16.png}
%  	\captionsetup{justification=centering}
%  	\caption{Errors by occlusion in skyscrapers}
% 	\label{fig:fig16}
% \end{figure*}

\subsubsection{Low point density scenarios} \label{s4.3.2}

Commission error of overhanging trees adjacent to buildings can occur more often with low point density scenarios. Note the orange pixel in Figure \ref{fig:fig14}. Since the laser penetrates the tree, the orange pixel will be classified as a non-building class, and it allows the separation between the building and the tree. However, if the tree is too dense to allow the laser penetration, the building and the tree will be merged into one object of the building candidate. If merged, it could either be removed or remain based on its planarity, which eventually results in either omission or commission error, respectively. Naturally, this error can occur more frequently if point density is low. 

\begin{figure}
	\centering
	\includegraphics[width=3.5in]{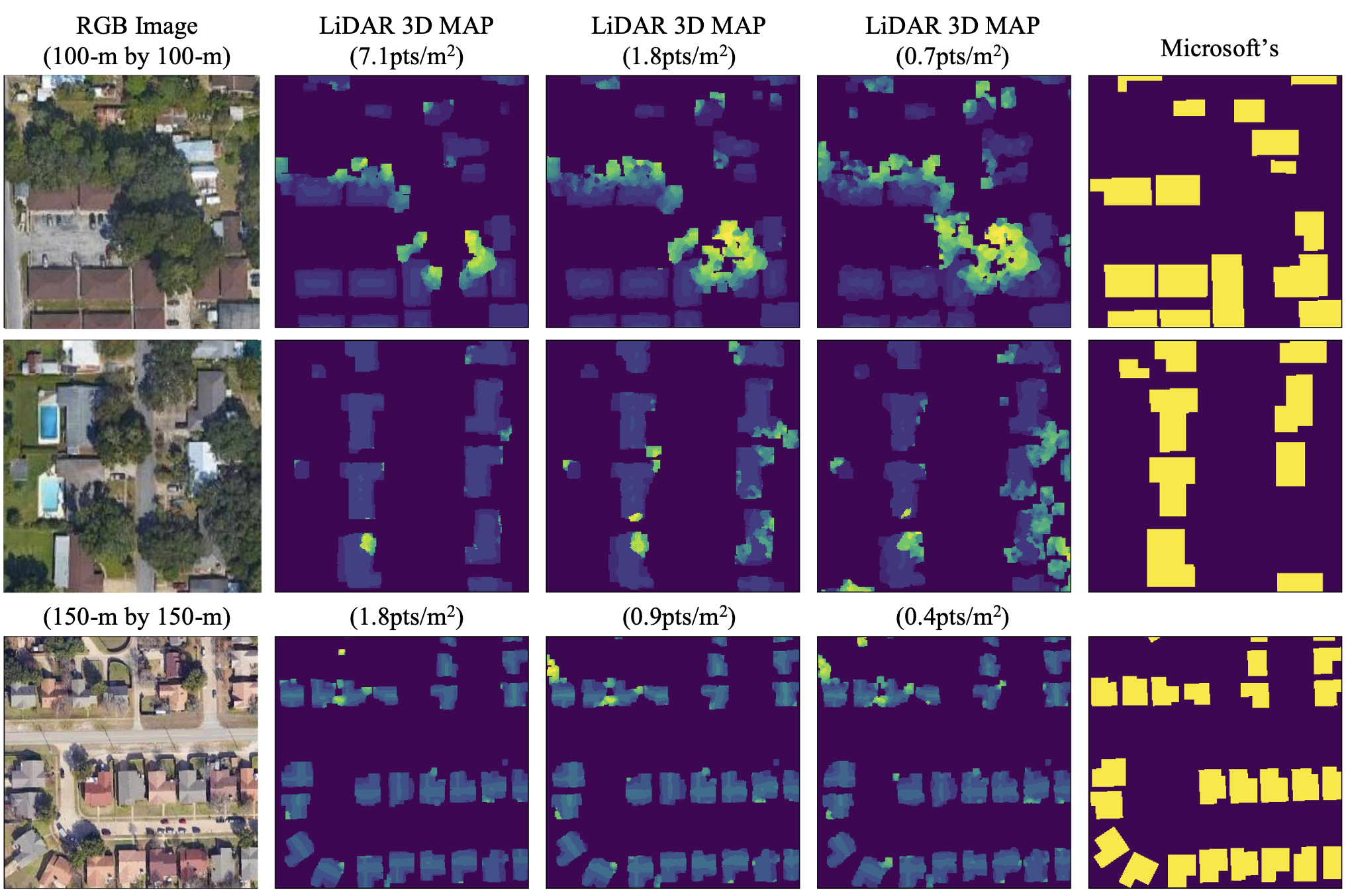}
%  	\captionsetup{justification=centering}
 	\caption{Errors due to overhanging trees in low point density scenarios}
	\label{fig:fig17}
     \vspace{-0.25in}
\end{figure}

To investigate the impact of the point density, we simulated low point density scenarios by subsampling the original LiDAR data. Figure \ref{fig:fig17} displays several challenging examples with overhanging dense trees. The first two rows are excerpted from Callaway, Florida, US, and the last row is excerpted from Dallas, Texas, US. LiDAR data were scanned from April to May of 2017 for the Callaway dataset and from April to July of 2019 for the Dallas dataset, respectively, which guarantees the season of fully leaf-on trees.
% A Riegl Q-1560 LiDAR system and a Leica ALS80 were used for the Callaway dataset and the Dallas dataset, respectively. RGB satellite images were from Google Earth. To visualize the overhanging trees effectively, LiDAR 3D maps were provided. Microsoft Building Footprints are also provided for reference. 

% \begin{figure*}
% 	\centering
% 	\includegraphics[width=6in]{figures/fig17.png}
%  	\captionsetup{justification=centering}
%  	\caption{Errors by overhanging dense trees in low point density scenarios}
% 	\label{fig:fig17}
% \end{figure*}

As the point density decreases, some trees are classified as buildings. Also, it results in an unreliable 3D building map when trees are over the building. Nevertheless, even with the very low point density scenario (<1-points/{m\textsuperscript{2}}), our workflow was able to properly detect buildings if they have no overhanging trees. This experiment shows that our workflow may require high point density LiDAR for generating reliable results, particularly for forested areas, but also suggests that the performance of our workflow will further be improved as point density increases.\newline

\subsubsection{Commission errors due to the DTM} \label{s4.3.3}

The robustness of the workflow can be attributed in part to the modification and integration of a well-suited DTM generation algorithm (i.e., DTM*). As DTM* classifies bridges and overpasses as terrain, it can prevent lots of errors that can occur if other DTM algorithms were embedded in our workflow as shown in Figure \ref{fig:fig1.5}. Nevertheless, some limitations associated with DTM were observed.

Errors can occur at the edge of the input LiDAR tile (the geographic boundaries of the input LiDAR data file). This is because DTM* assumes small, non-rectangular areas enclosed by a certain level of a steep slope as non-grounds, but the enclosed area near the edge of the LiDAR could be a disconnected terrain (please refer to \cite{song2022dtm} for the detailed description). Likewise, some terrains surrounded by steep slopes could be extracted as building.

\begin{figure}
	\centering
	\includegraphics[width=3.5in]{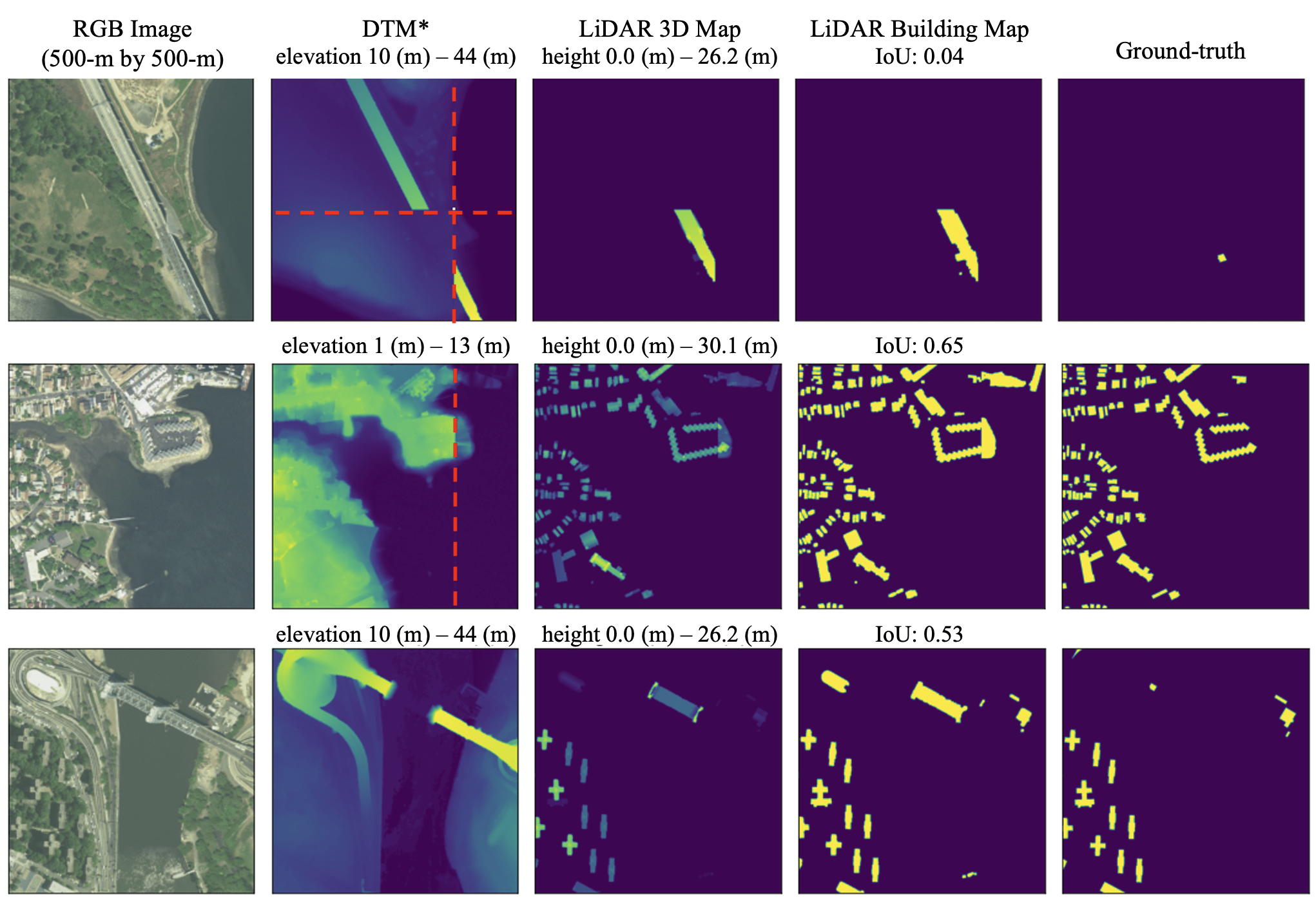}
%  	\captionsetup{justification=centering}
 	\caption{Commission errors caused by artifacts in DTM (The red dashed line indicates the data boundary of the LiDAR tile)}
	\label{fig:fig18}
\end{figure}

Figure \ref{fig:fig18} provides some errors caused by artifacts of DTM. The red dashed line delineates the boundary of LiDAR tiles. The first row shows the region where four different LiDAR tiles are adjacent to each other. Since the bridge in the lower-left LiDAR tile was disconnected, the bridge was classified as a non-ground object by DTM*, and as a result, the disconnected part was classified as a building in our workflow. The second row shows another example of an error caused by a DTM error occurring at the edge of the LiDAR tiles. Small disconnected lands near the shoreline were classified as buildings. The third row shows an example of a suspension bridge. As pylons of the suspension bridge disconnected the two ends of the bridge, the middle part of the bridge was classified as a building.

% \begin{figure*}
% 	\centering
% 	\includegraphics[width=6in]{figures/fig18.png}
%  	\captionsetup{justification=centering}
%  	\caption{Commission errors caused by artifacts in DTM (The red dashed line indicates the data boundary of the LiDAR tile)}
% 	\label{fig:fig18}
% \end{figure*}

Errors occurring at the edge of the LiDAR tiles can be prevented by simply using a larger extent of input for data processing. Also, when splitting the entire point clouds into multiple tiles is needed due to limited computational resources, employing our workflow with an overlapping sliding window and outputting the central part of each window would prevent errors from DTM.\newline

\subsubsection{Deformation during the morphological filtering} \label{s4.3.4}

Our workflow includes the process that extracts buildings from among building candidates using morphological operations. One problem is that dilation after erosion may not recover the original shape of the object. If a rectangular building’s edge is not aligned with the X or Y axis, the corner can be blunted, which ultimately changes the building area after the operation.

\begin{figure}
	\centering
	\includegraphics[width=3.5in]{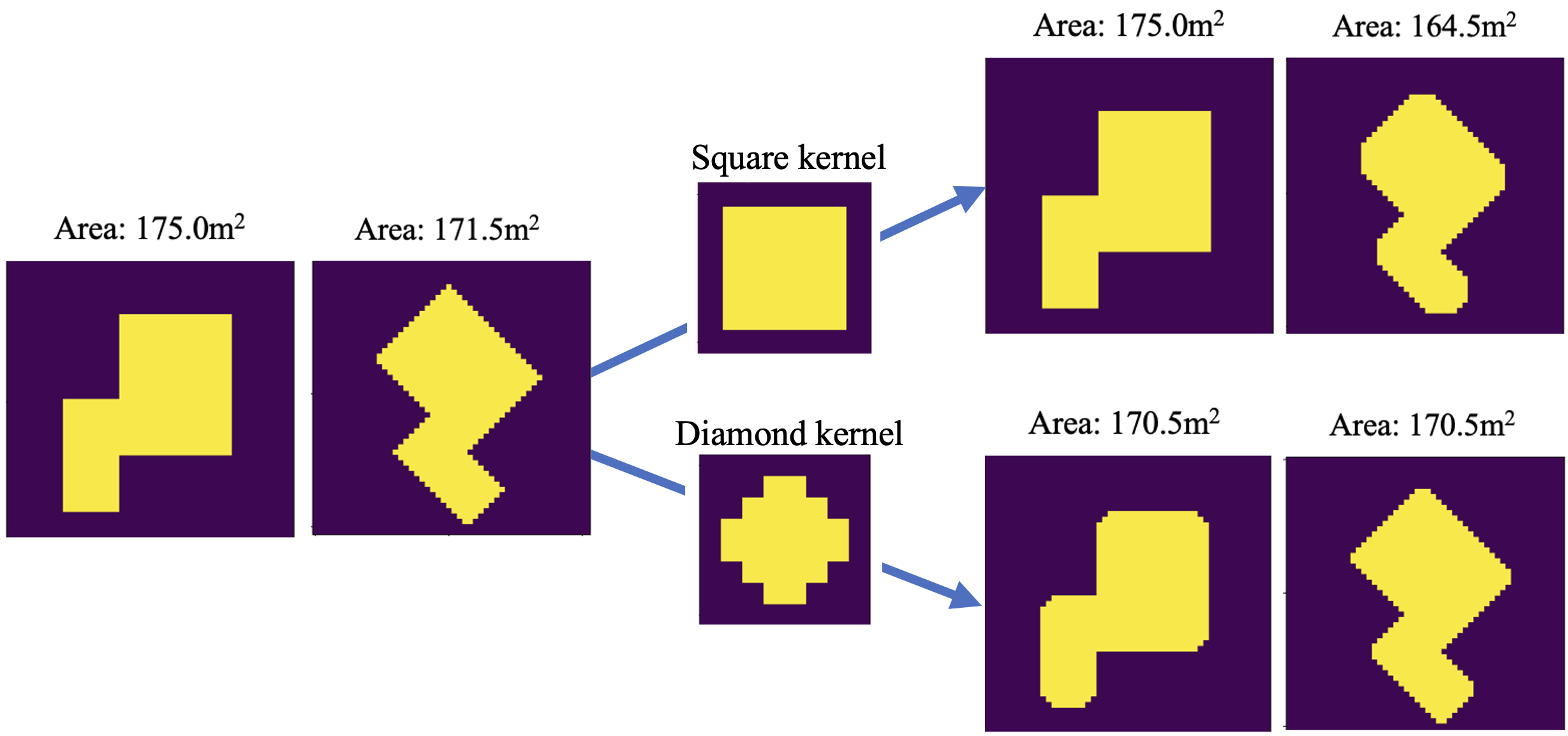}
	%\captionsetup{justification=centering}
	\caption{Deformations caused by morphological filters}
	\label{fig:fig19}
    \vspace{-0.2in}
\end{figure}

One alternative option to mitigate this deformation is to use a diamond-shaped kernel. Figure \ref{fig:fig19} illustrates a toy example of the morphological filtering process on the mockup residential buildings. It can alleviate the deformation, but it still cannot perfectly recover the original shape. The default K1, a square 7 by 7, produced reasonable accuracy in diverse study areas for the 0.5-meter by 0.5-meter resolution. However, if the focus is on modeling detailed building boundaries, other boundary-regularizing algorithms are recommended \cite{dos2019extraction, dos2020regularization, kamra2022lightweight}.\newline

% \begin{figure*}
% 	\centering
% 	\includegraphics[width=6in]{figures/fig19.png}
% 	\captionsetup{justification=centering}
% 	\caption{Examples of deformations caused by morphological filters}
% 	\label{fig:fig19}
% \end{figure*}
% \subsubsection{Limitations in 3D building map} \label{s4.3.5}

\subsubsection{Limitations in 3D building map} \label{s4.3.5}

While our workflow generally produces reliable 3D building maps, we have identified some inaccuracies near high-rise buildings and residential buildings obscured by overhanging trees, as shown in Figure \ref{fig:fig16} and Figure \ref{fig:fig17}. Although applying a median filter for roof smoothing enhances the model, it can potentially erase minor sub-structures like chimneys due to over-regularization. The quality of the model is dependent on the LiDAR data's precision, and additional geometric evaluation might require more accurate reference measurements. It's also important to note that our 3D building map may not capture the fine details of buildings accurately, as the primary focus of our study is on scalable mass production of rasterized building maps. The map does not portray the interior or sides of buildings, and color information is not provided. These are inherent limitations that users should be aware of when employing our workflow. Future work is needed to refine and enhance the details of 3D building models \cite{kamra2022lightweight,wang2023reconstruction} and to develop a method for assessing the quality of these models over large areas.

\section{Conclusion} \label{s5.}
We present an end-to-end open-source workflow for 2D and 3D building mapping from raw airborne LiDAR data. To our knowledge, there has been no well-established open-source workflow for generating 2D and 3D building maps. Our system operates fully unsupervised, is computationally efficient, and produces accurate results without the need for intensive parameter tuning. Moreover, this study is the first to evaluate a LiDAR-based building map against deep learning-based and hand-digitized products on a large scale (> 550 {km\textsuperscript{2}}). Our work not only provides a practical solution for mass producing 2D and 3D building maps, but also furnishes valuable insights into the strengths and weaknesses of different methodologies, informing expectations for major large-area building mapping techniques.

In recent years, deep learning-based methods, bolstered by advancements in deep learning and the ready availability of image data, have dominated the literature. However, such methods have notable drawbacks. They typically require training procedures, are computationally expensive, and their errors can be unpredictable and difficult to explain. While deep learning-based methods can achieve high accuracy according to quantitative metrics, if users cannot anticipate or understand the errors or biases these models may introduce, subsequent studies based on these maps may lead to skewed outcomes.

Our workflow, on the other hand, is based on a simple yet robust assumption that a building is a ground-standing object with a relatively smooth, laser-impermeable surface. As such, our workflow is highly scalable and its results are readily explainable. Moreover, since the operation is completely unsupervised and straightforward, all procedures are transparent, unlike the ``black box'' nature of deep learning.

Despite its outperforming performance, our workflow does have certain limitations. Most of these can be mitigated with better quality (higher point density) LiDAR data. The remaining challenges include the unclear boundaries between dense trees and small buildings, the misclassification of non-building structures as buildings, deformation during morphological filtering, and the low level of detail in building models. These issues necessitate further research. If the study area is small, adding more sophisticated rules to our workflow may fix some errors. However, if the goal is a large-area mapping where all outputs cannot be validated after the mapping, careful consideration is needed not to lose the generalization performance.

The source code of our workflow will be publicly available via GitHub \emph{(by the date of publication)}. Also, we will continue to refine the open-source code, and the generated 2D and 3D building maps will be released and kept updated. Our open-source workflow will enable the community to generate 2D and 3D building maps more readily and accurately, and we hope our workflow and findings contribute to supporting various studies with better building maps.

\section*{Acknowledgement}
All LiDAR data used in this paper are from the U.S. Geological Survey’s 3D Elevation Program (3DEP).% We would like to thank everyone who has put in a tremendous amount of effort for 3DEP.

% {\appendix[Proof of the Zonklar Equations]
% Use $\backslash${\tt{appendix}} if you have a single appendix:
% Do not use $\backslash${\tt{section}} anymore after $\backslash${\tt{appendix}}, only $\backslash${\tt{section*}}.
% If you have multiple appendixes use $\backslash${\tt{appendices}} then use $\backslash${\tt{section}} to start each appendix.
% You must declare a $\backslash${\tt{section}} before using any $\backslash${\tt{subsection}} or using $\backslash${\tt{label}} ($\backslash${\tt{appendices}} by itself
%  starts a section numbered zero.)}

% %{\appendices
% %\section*{Proof of the First Zonklar Equation}
% %Appendix one text goes here.
% % You can choose not to have a title for an appendix if you want by leaving the argument blank
% %\section*{Proof of the Second Zonklar Equation}
% %Appendix two text goes here.}

% \section{References Section}
% You can use a bibliography generated by BibTeX as a .bbl file.
%  BibTeX documentation can be easily obtained at:
%  http://mirror.ctan.org/biblio/bibtex/contrib/doc/
%  The IEEEtran BibTeX style support page is:
%  http://www.michaelshell.org/tex/ieeetran/bibtex/
 
%  % argument is your BibTeX string definitions and bibliography database(s)
% %\bibliography{IEEEabrv,../bib/paper}
% You can manually copy in the resultant .bbl file and set second argument of $\backslash${\tt{begin}} to the number of references
%  (used to reserve space for the reference number labels box).

\bibliographystyle{IEEEtran}
\bibliography{references}

\begin{thebibliography}{10}
\providecommand{\url}[1]{#1}
\csname url@samestyle\endcsname
\providecommand{\newblock}{\relax}
\providecommand{\bibinfo}[2]{#2}
\providecommand{\BIBentrySTDinterwordspacing}{\spaceskip=0pt\relax}
\providecommand{\BIBentryALTinterwordstretchfactor}{4}
\providecommand{\BIBentryALTinterwordspacing}{\spaceskip=\fontdimen2\font plus
\BIBentryALTinterwordstretchfactor\fontdimen3\font minus
  \fontdimen4\font\relax}
\providecommand{\BIBforeignlanguage}[2]{{%
\expandafter\ifx\csname l@#1\endcsname\relax
\typeout{** WARNING: IEEEtran.bst: No hyphenation pattern has been}%
\typeout{** loaded for the language `#1'. Using the pattern for}%
\typeout{** the default language instead.}%
\else
\language=\csname l@#1\endcsname
\fi
#2}}
\providecommand{\BIBdecl}{\relax}
\BIBdecl

\bibitem{zhu2019understanding}
Z.~Zhu, Y.~Zhou, K.~C. Seto, E.~C. Stokes, C.~Deng, S.~T. Pickett, and
  H.~Taubenb{\"o}ck, ``Understanding an urbanizing planet: Strategic directions
  for remote sensing,'' \emph{Remote Sensing of Environment}, vol. 228, pp.
  164--182, 2019.

\bibitem{ghaffarian2019post}
S.~Ghaffarian, N.~Kerle, E.~Pasolli, and J.~Jokar~Arsanjani, ``Post-disaster
  building database updating using automated deep learning: An integration of
  pre-disaster openstreetmap and multi-temporal satellite data,'' \emph{Remote
  sensing}, vol.~11, no.~20, p. 2427, 2019.

\bibitem{hong2019temporal}
J.-W. Hong, J.~Hong, E.~E. Kwon, and D.~Yoon, ``Temporal dynamics of urban heat
  island correlated with the socio-economic development over the past
  half-century in seoul, korea,'' \emph{Environmental Pollution}, vol. 254, p.
  112934, 2019.

\bibitem{herbert2015comparison}
G.~Herbert and X.~Chen, ``A comparison of usefulness of 2d and 3d
  representations of urban planning,'' \emph{Cartography and Geographic
  Information Science}, vol.~42, no.~1, pp. 22--32, 2015.

\bibitem{wu2005population}
S.-s. Wu, X.~Qiu, and L.~Wang, ``Population estimation methods in gis and
  remote sensing: A review,'' \emph{GIScience \& Remote Sensing}, vol.~42,
  no.~1, pp. 80--96, 2005.

\bibitem{herfort2021evolution}
B.~Herfort, S.~Lautenbach, J.~Porto~de Albuquerque, J.~Anderson, and A.~Zipf,
  ``The evolution of humanitarian mapping within the openstreetmap community,''
  \emph{Scientific reports}, vol.~11, no.~1, pp. 1--15, 2021.

\bibitem{harig2021automatic}
O.~Harig, R.~Hecht, D.~Burghardt, and G.~Meinel, ``Automatic delineation of
  urban growth boundaries based on topographic data using germany as a case
  study,'' \emph{ISPRS International Journal of Geo-Information}, vol.~10,
  no.~5, p. 353, 2021.

\bibitem{hecht2013measuring}
R.~Hecht, C.~Kunze, and S.~Hahmann, ``Measuring completeness of building
  footprints in openstreetmap over space and time,'' \emph{ISPRS International
  Journal of Geo-Information}, vol.~2, no.~4, pp. 1066--1091, 2013.

\bibitem{vargas2020openstreetmap}
J.~E. Vargas-Munoz, S.~Srivastava, D.~Tuia, and A.~X. Falcao, ``Openstreetmap:
  Challenges and opportunities in machine learning and remote sensing,''
  \emph{IEEE Geoscience and Remote Sensing Magazine}, vol.~9, no.~1, pp.
  184--199, 2020.

\bibitem{dorn2015quality}
H.~Dorn, T.~T{\"o}rnros, and A.~Zipf, ``Quality evaluation of vgi using
  authoritative data—a comparison with land use data in southern germany,''
  \emph{ISPRS International Journal of Geo-Information}, vol.~4, no.~3, pp.
  1657--1671, 2015.

\bibitem{jokar2015quality}
J.~Jokar~Arsanjani, P.~Mooney, A.~Zipf, and A.~Schauss, ``Quality assessment of
  the contributed land use information from openstreetmap versus authoritative
  datasets,'' in \emph{OpenStreetMap in GIScience}.\hskip 1em plus 0.5em minus
  0.4em\relax Springer, 2015, pp. 37--58.

\bibitem{van2018spacenet}
A.~Van~Etten, D.~Lindenbaum, and T.~M. Bacastow, ``Spacenet: A remote sensing
  dataset and challenge series,'' \emph{arXiv preprint arXiv:1807.01232}, 2018.

\bibitem{huang2023building}
Z.~Huang, Q.~Liu, H.~Zhou, G.~Gao, T.~Xu, Q.~Wen, and Y.~Wang, ``Building
  detection from panchromatic and multispectral images with dual-stream
  asymmetric fusion networks,'' \emph{IEEE Journal of Selected Topics in
  Applied Earth Observations and Remote Sensing}, vol.~16, pp. 3364--3377,
  2023.

\bibitem{ma2019deep}
L.~Ma, Y.~Liu, X.~Zhang, Y.~Ye, G.~Yin, and B.~A. Johnson, ``Deep learning in
  remote sensing applications: A meta-analysis and review,'' \emph{ISPRS
  journal of photogrammetry and remote sensing}, vol. 152, pp. 166--177, 2019.

\bibitem{zhu2017deep}
X.~X. Zhu, D.~Tuia, L.~Mou, G.-S. Xia, L.~Zhang, F.~Xu, and F.~Fraundorfer,
  ``Deep learning in remote sensing: A comprehensive review and list of
  resources,'' \emph{IEEE Geoscience and Remote Sensing Magazine}, vol.~5,
  no.~4, pp. 8--36, 2017.

\bibitem{yang2018building}
H.~L. Yang, J.~Yuan, D.~Lunga, M.~Laverdiere, A.~Rose, and B.~Bhaduri,
  ``Building extraction at scale using convolutional neural network: Mapping of
  the united states,'' \emph{IEEE Journal of Selected Topics in Applied Earth
  Observations and Remote Sensing}, vol.~11, no.~8, pp. 2600--2614, 2018.

\bibitem{maggiori2016convolutional}
E.~Maggiori, Y.~Tarabalka, G.~Charpiat, and P.~Alliez, ``Convolutional neural
  networks for large-scale remote-sensing image classification,'' \emph{IEEE
  Transactions on geoscience and remote sensing}, vol.~55, no.~2, pp. 645--657,
  2016.

\bibitem{deng2019large}
X.~Deng, H.~L. Yang, N.~Makkar, and D.~Lunga, ``Large scale unsupervised domain
  adaptation of segmentation networks with adversarial learning,'' in
  \emph{IGARSS 2019-2019 IEEE International Geoscience and Remote Sensing
  Symposium}.\hskip 1em plus 0.5em minus 0.4em\relax IEEE, 2019, pp.
  4955--4958.

\bibitem{makkar2021adversarial}
N.~Makkar, L.~Yang, and S.~Prasad, ``Adversarial learning based discriminative
  domain adaptation for geospatial image analysis,'' \emph{IEEE Journal of
  Selected Topics in Applied Earth Observations and Remote Sensing}, vol.~15,
  pp. 150--162, 2021.

\bibitem{dias2022model}
P.~Dias, Y.~Tian, S.~Newsam, A.~Tsaris, J.~Hinkle, and D.~Lunga, ``Model
  assumptions and data characteristics: Impacts on domain adaptation in
  building segmentation,'' \emph{IEEE Transactions on Geoscience and Remote
  Sensing}, vol.~60, pp. 1--18, 2022.

\bibitem{MS_BUILDING}
``{Microsoft Building Footprints},''
  \url{https://www.microsoft.com/en-us/maps/building-footprints}, accessed:
  2022-09-27.

\bibitem{tan2019efficientnet}
M.~Tan and Q.~Le, ``Efficientnet: Rethinking model scaling for convolutional
  neural networks,'' in \emph{International conference on machine
  learning}.\hskip 1em plus 0.5em minus 0.4em\relax PMLR, 2019, pp. 6105--6114.

\bibitem{heris2020rasterized}
M.~P. Heris, N.~L. Foks, K.~J. Bagstad, A.~Troy, and Z.~H. Ancona, ``A
  rasterized building footprint dataset for the united states,''
  \emph{Scientific data}, vol.~7, no.~1, pp. 1--10, 2020.

\bibitem{williams2019mapping}
T.~K.-A. Williams, T.~Wei, and X.~Zhu, ``Mapping urban slum settlements using
  very high-resolution imagery and land boundary data,'' \emph{IEEE Journal of
  Selected Topics in Applied Earth Observations and Remote Sensing}, vol.~13,
  pp. 166--177, 2019.

\bibitem{li2020continental}
M.~Li, E.~Koks, H.~Taubenb{\"o}ck, and J.~van Vliet, ``Continental-scale
  mapping and analysis of 3d building structure,'' \emph{Remote Sensing of
  Environment}, vol. 245, p. 111859, 2020.

\bibitem{park2021impacts}
Y.~Park, J.-M. Guldmann, and D.~Liu, ``Impacts of tree and building shades on
  the urban heat island: Combining remote sensing, 3d digital city and spatial
  regression approaches,'' \emph{Computers, Environment and Urban Systems},
  vol.~88, p. 101655, 2021.

\bibitem{macchione2019moving}
F.~Macchione, P.~Costabile, C.~Costanzo, and R.~De~Santis, ``Moving to 3-d
  flood hazard maps for enhancing risk communication,'' \emph{Environmental
  modelling \& software}, vol. 111, pp. 510--522, 2019.

\bibitem{wang2016fine}
S.~Wang, Y.~Tian, Y.~Zhou, W.~Liu, and C.~Lin, ``Fine-scale population
  estimation by 3d reconstruction of urban residential buildings,''
  \emph{Sensors}, vol.~16, no.~10, p. 1755, 2016.

\bibitem{biljecki2016population}
F.~Biljecki, K.~Arroyo~Ohori, H.~Ledoux, R.~Peters, and J.~Stoter, ``Population
  estimation using a 3d city model: A multi-scale country-wide study in the
  netherlands,'' \emph{PloS one}, vol.~11, no.~6, p. e0156808, 2016.

\bibitem{biljecki2016variants}
F.~Biljecki, H.~Ledoux, J.~Stoter, and G.~Vosselman, ``The variants of an lod
  of a 3d building model and their influence on spatial analyses,'' \emph{ISPRS
  Journal of Photogrammetry and Remote Sensing}, vol. 116, pp. 42--54, 2016.

\bibitem{kim2021regionalization}
N.~Kim and Y.~Yoon, ``Regionalization for urban air mobility application with
  analyses of 3d urban space and geodemography in san francisco and new york,''
  \emph{Procedia Computer Science}, vol. 184, pp. 388--395, 2021.

\bibitem{lehner2020digital}
H.~Lehner and L.~Dorffner, ``Digital geotwin vienna: towards a digital twin
  city as geodata hub,'' 2020.

\bibitem{han2022utilising}
J.-Y. Han, Y.-C. Chen, and S.-Y. Li, ``Utilising high-fidelity 3d building
  model for analysing the rooftop solar photovoltaic potential in urban
  areas,'' \emph{Solar Energy}, vol. 235, pp. 187--199, 2022.

\bibitem{zhou2019community}
Z.~Zhou, J.~Gong, and X.~Hu, ``Community-scale multi-level post-hurricane
  damage assessment of residential buildings using multi-temporal airborne
  lidar data,'' \emph{Automation in Construction}, vol.~98, pp. 30--45, 2019.

\bibitem{kamra2022lightweight}
V.~Kamra, P.~Kudeshia, S.~ArabiNaree, D.~Chen, Y.~Akiyama, and J.~Peethambaran,
  ``Lightweight reconstruction of urban buildings: Data structures, algorithms,
  and future directions,'' \emph{IEEE Journal of Selected Topics in Applied
  Earth Observations and Remote Sensing}, vol.~16, pp. 902--917, 2022.

\bibitem{qin2019critical}
R.~Qin, ``A critical analysis of satellite stereo pairs for digital surface
  model generation and a matching quality prediction model,'' \emph{ISPRS
  Journal of Photogrammetry and Remote Sensing}, vol. 154, pp. 139--150, 2019.

\bibitem{huang2022evaluation}
D.~Huang, Y.~Tang, and R.~Qin, ``An evaluation of planetscope images for 3d
  reconstruction and change detection--experimental validations with case
  studies,'' \emph{GIScience \& Remote Sensing}, vol.~59, no.~1, pp. 744--761,
  2022.

\bibitem{dominguez2019back}
E.~M. Dom{\'\i}nguez, C.~Magnard, E.~Meier, D.~Small, M.~E. Schaepman, and
  D.~Henke, ``A back-projection tomographic framework for vhr sar image change
  detection,'' \emph{IEEE Transactions on Geoscience and Remote Sensing},
  vol.~57, no.~7, pp. 4470--4484, 2019.

\bibitem{rottensteiner2012isprs}
F.~Rottensteiner, G.~Sohn, J.~Jung, M.~Gerke, C.~Baillard, S.~Benitez, and
  U.~Breitkopf, ``The isprs benchmark on urban object classification and 3d
  building reconstruction,'' \emph{ISPRS Annals of the Photogrammetry, Remote
  Sensing and Spatial Information Sciences I-3 (2012), Nr. 1}, vol.~1, no.~1,
  pp. 293--298, 2012.

\bibitem{rottensteiner2014results}
F.~Rottensteiner, G.~Sohn, M.~Gerke, J.~D. Wegner, U.~Breitkopf, and J.~Jung,
  ``Results of the isprs benchmark on urban object detection and 3d building
  reconstruction,'' \emph{ISPRS journal of photogrammetry and remote sensing},
  vol.~93, pp. 256--271, 2014.

\bibitem{tuia2016domain}
D.~Tuia, C.~Persello, and L.~Bruzzone, ``Domain adaptation for the
  classification of remote sensing data: An overview of recent advances,''
  \emph{IEEE geoscience and remote sensing magazine}, vol.~4, no.~2, pp.
  41--57, 2016.

\bibitem{wang2023automatic}
F.~Wang, G.~Zhou, J.~Xie, B.~Fu, H.~You, J.~Chen, X.~Shi, and B.~Zhou, ``An
  automatic hierarchical clustering method for the lidar point cloud
  segmentation of buildings via shape classification and outliers
  reassignment,'' \emph{Remote Sensing}, vol.~15, no.~9, p. 2432, 2023.

\bibitem{liu2023roof}
K.~Liu, H.~Ma, L.~Zhang, X.~Liang, D.~Chen, and Y.~Liu, ``Roof segmentation
  from airborne lidar using octree-based hybrid region growing and boundary
  neighborhood verification voting,'' \emph{IEEE Journal of Selected Topics in
  Applied Earth Observations and Remote Sensing}, vol.~16, pp. 2134--2146,
  2023.

\bibitem{wang2023reconstruction}
F.~Wang, G.~Zhou, H.~Hu, Y.~Wang, B.~Fu, S.~Li, and J.~Xie, ``Reconstruction of
  lod-2 building models guided by fa{\c{c}}ade structures from oblique
  photogrammetric point cloud,'' \emph{Remote Sensing}, vol.~15, no.~2, p. 400,
  2023.

\bibitem{huang2022city3d}
J.~Huang, J.~Stoter, R.~Peters, and L.~Nan, ``City3d: Large-scale building
  reconstruction from airborne lidar point clouds,'' \emph{Remote Sensing},
  vol.~14, no.~9, p. 2254, 2022.

\bibitem{lewandowicz20223d}
E.~Lewandowicz, F.~Tarsha~Kurdi, and Z.~Gharineiat, ``3d lod2 and lod3 modeling
  of buildings with ornamental towers and turrets based on lidar data,''
  \emph{Remote Sensing}, vol.~14, no.~19, p. 4687, 2022.

\bibitem{li2022recursive}
X.~Li, F.~Qiu, F.~Shi, and Y.~Tang, ``A recursive hull and signal-based
  building footprint generation from airborne lidar data,'' \emph{Remote
  Sensing}, vol.~14, no.~22, p. 5892, 2022.

\bibitem{bizjak2023novel}
M.~Bizjak, D.~Mongus, B.~{\v{Z}}alik, and N.~Luka{\v{c}}, ``Novel half-spaces
  based 3d building reconstruction using airborne lidar data,'' \emph{Remote
  Sensing}, vol.~15, no.~5, p. 1269, 2023.

\bibitem{dey2023machine}
E.~K. Dey, M.~Awrangjeb, F.~Tarsha~Kurdi, and B.~Stantic, ``Machine
  learning-based segmentation of aerial lidar point cloud data on building
  roof,'' \emph{European Journal of Remote Sensing}, vol.~56, no.~1, p.
  2210745, 2023.

\bibitem{li2022ransac}
Z.~Li and J.~Shan, ``Ransac-based multi primitive building reconstruction from
  3d point clouds,'' \emph{ISPRS Journal of Photogrammetry and Remote Sensing},
  vol. 185, pp. 247--260, 2022.

\bibitem{zhang2021optimal}
W.~Zhang, Z.~Li, and J.~Shan, ``Optimal model fitting for building
  reconstruction from point clouds,'' \emph{IEEE Journal of Selected Topics in
  Applied Earth Observations and Remote Sensing}, vol.~14, pp. 9636--9650,
  2021.

\bibitem{li2022point2roof}
L.~Li, N.~Song, F.~Sun, X.~Liu, R.~Wang, J.~Yao, and S.~Cao, ``Point2roof:
  End-to-end 3d building roof modeling from airborne lidar point clouds,''
  \emph{ISPRS Journal of Photogrammetry and Remote Sensing}, vol. 193, pp.
  17--28, 2022.

\bibitem{tarsha2022automatic}
F.~Tarsha~Kurdi, Z.~Gharineiat, G.~Campbell, M.~Awrangjeb, and E.~K. Dey,
  ``Automatic filtering of lidar building point cloud in case of trees
  associated to building roof,'' \emph{Remote Sensing}, vol.~14, no.~2, p. 430,
  2022.

\bibitem{groger2012citygml}
G.~Gr{\"o}ger and L.~Pl{\"u}mer, ``Citygml--interoperable semantic 3d city
  models,'' \emph{ISPRS Journal of Photogrammetry and Remote Sensing}, vol.~71,
  pp. 12--33, 2012.

\bibitem{volk2014building}
R.~Volk, J.~Stengel, and F.~Schultmann, ``Building information modeling (bim)
  for existing buildings—literature review and future needs,''
  \emph{Automation in construction}, vol.~38, pp. 109--127, 2014.

\bibitem{hyyppa2001segmentation}
J.~Hyyppa, O.~Kelle, M.~Lehikoinen, and M.~Inkinen, ``A segmentation-based
  method to retrieve stem volume estimates from 3-d tree height models produced
  by laser scanners,'' \emph{IEEE Transactions on geoscience and remote
  sensing}, vol.~39, no.~5, pp. 969--975, 2001.

\bibitem{jung2014framework}
J.~Jung, E.~Pasolli, S.~Prasad, J.~C. Tilton, and M.~M. Crawford, ``A framework
  for land cover classification using discrete return lidar data: Adopting
  pseudo-waveform and hierarchical segmentation,'' \emph{IEEE Journal of
  Selected Topics in Applied Earth Observations and Remote Sensing}, vol.~7,
  no.~2, pp. 491--502, 2014.

\bibitem{maltezos2018building}
E.~Maltezos, A.~Doulamis, N.~Doulamis, and C.~Ioannidis, ``Building extraction
  from lidar data applying deep convolutional neural networks,'' \emph{IEEE
  Geoscience and Remote Sensing Letters}, vol.~16, no.~1, pp. 155--159, 2018.

\bibitem{oh2022high}
S.~Oh, J.~Jung, G.~Shao, G.~Shao, J.~Gallion, and S.~Fei, ``High-resolution
  canopy height model generation and validation using usgs 3dep lidar data in
  indiana, usa,'' \emph{Remote Sensing}, vol.~14, no.~4, p. 935, 2022.

\bibitem{awrangjeb2012building}
M.~Awrangjeb, C.~Zhang, C.~S. Fraser \emph{et~al.}, ``Building detection in
  complex scenes thorough effective separation of buildings from trees,''
  \emph{Photogrammetric Engineering \& Remote Sensing}, vol.~78, no.~7, pp.
  729--745, 2012.

\bibitem{huang2019automatic}
J.~Huang, X.~Zhang, Q.~Xin, Y.~Sun, and P.~Zhang, ``Automatic building
  extraction from high-resolution aerial images and lidar data using gated
  residual refinement network,'' \emph{ISPRS journal of photogrammetry and
  remote sensing}, vol. 151, pp. 91--105, 2019.

\bibitem{chen2020automatic}
S.~Chen, W.~Shi, M.~Zhou, M.~Zhang, and P.~Chen, ``Automatic building
  extraction via adaptive iterative segmentation with lidar data and high
  spatial resolution imagery fusion,'' \emph{IEEE Journal of Selected Topics in
  Applied Earth Observations and Remote Sensing}, vol.~13, pp. 2081--2095,
  2020.

\bibitem{zhao2016extracting}
Z.~Zhao, Y.~Duan, Y.~Zhang, and R.~Cao, ``Extracting buildings from and
  regularizing boundaries in airborne lidar data using connected operators,''
  \emph{International Journal of Remote Sensing}, vol.~37, no.~4, pp. 889--912,
  2016.

\bibitem{yan2017hierarchical}
Y.~Yan, F.~Gao, S.~Deng, and N.~Su, ``A hierarchical building segmentation in
  digital surface models for 3d reconstruction,'' \emph{Sensors}, vol.~17,
  no.~2, p. 222, 2017.

\bibitem{yuan2021multiscale}
Q.~Yuan, H.~Z.~M. Shafri, A.~H. Alias, and S.~J.~b. Hashim, ``Multiscale
  semantic feature optimization and fusion network for building extraction
  using high-resolution aerial images and lidar data,'' \emph{Remote Sensing},
  vol.~13, no.~13, p. 2473, 2021.

\bibitem{hosseinpour2022cmgfnet}
H.~Hosseinpour, F.~Samadzadegan, and F.~D. Javan, ``Cmgfnet: A deep cross-modal
  gated fusion network for building extraction from very high-resolution remote
  sensing images,'' \emph{ISPRS journal of photogrammetry and remote sensing},
  vol. 184, pp. 96--115, 2022.

\bibitem{ojogbane2021automated}
S.~S. Ojogbane, S.~Mansor, B.~Kalantar, Z.~B. Khuzaimah, H.~Z.~M. Shafri, and
  N.~Ueda, ``Automated building detection from airborne lidar and very
  high-resolution aerial imagery with deep neural network,'' \emph{Remote
  Sensing}, vol.~13, no.~23, p. 4803, 2021.

\bibitem{morgan2002interpolation}
M.~Morgan and A.~Habib, ``Interpolation of lidar data and automatic building
  extraction,'' in \emph{ACSM-ASPRS Annual conference proceedings}.\hskip 1em
  plus 0.5em minus 0.4em\relax Citeseer, 2002, pp. 432--441.

\bibitem{song2022dtm}
H.~Song and J.~Jung, ``A new explainable dtm generation algorithm with airborne
  lidar data: grounds are smoothly connected eventually,'' \emph{arXiv preprint
  arXiv:2208.11243}, 2022.

\bibitem{sithole2003report}
G.~Sithole and G.~Vosselman, ``Report: Isprs comparison of filters,''
  \emph{ISPRS commission III, working group}, vol.~3, 2003.

\bibitem{meng2010ground}
X.~Meng, N.~Currit, and K.~Zhao, ``Ground filtering algorithms for airborne
  lidar data: A review of critical issues,'' \emph{Remote Sensing}, vol.~2,
  no.~3, pp. 833--860, 2010.

\bibitem{hofle2009water}
B.~H{\"o}fle, M.~Vetter, N.~Pfeifer, G.~Mandlburger, and J.~St{\"o}tter,
  ``Water surface mapping from airborne laser scanning using signal intensity
  and elevation data,'' \emph{Earth Surface Processes and Landforms}, vol.~34,
  no.~12, pp. 1635--1649, 2009.

\bibitem{liu2013automatic}
C.~Liu, B.~Shi, X.~Yang, N.~Li, and H.~Wu, ``Automatic buildings extraction
  from lidar data in urban area by neural oscillator network of visual
  cortex,'' \emph{IEEE Journal of Selected Topics in Applied Earth Observations
  and Remote Sensing}, vol.~6, no.~4, pp. 2008--2019, 2013.

\bibitem{du2017automatic}
S.~Du, Y.~Zhang, Z.~Zou, S.~Xu, X.~He, and S.~Chen, ``Automatic building
  extraction from lidar data fusion of point and grid-based features,''
  \emph{ISPRS journal of photogrammetry and remote sensing}, vol. 130, pp.
  294--307, 2017.

\bibitem{niemeyer2014contextual}
J.~Niemeyer, F.~Rottensteiner, and U.~Soergel, ``Contextual classification of
  lidar data and building object detection in urban areas,'' \emph{ISPRS
  journal of photogrammetry and remote sensing}, vol.~87, pp. 152--165, 2014.

\bibitem{rottensteiner2003automatic}
F.~Rottensteiner, ``Automatic generation of high-quality building models from
  lidar data,'' \emph{IEEE Computer Graphics and Applications}, vol.~23, no.~6,
  pp. 42--50, 2003.

\bibitem{yu2010automated}
B.~Yu, H.~Liu, J.~Wu, Y.~Hu, and L.~Zhang, ``Automated derivation of urban
  building density information using airborne lidar data and object-based
  method,'' \emph{Landscape and Urban Planning}, vol.~98, no. 3-4, pp.
  210--219, 2010.

\bibitem{dos2019extraction}
R.~C. dos Santos, M.~Galo, and A.~C. Carrilho, ``Extraction of building roof
  boundaries from lidar data using an adaptive alpha-shape algorithm,''
  \emph{IEEE Geoscience and Remote Sensing Letters}, vol.~16, no.~8, pp.
  1289--1293, 2019.

\bibitem{dos2022weighted}
R.~C. Dos~Santos, A.~F. Habib, and M.~Galo, ``Weighted iterative cd-spline for
  mitigating occlusion effects on building boundary regularization using
  airborne lidar data,'' \emph{Sensors}, vol.~22, no.~17, p. 6440, 2022.

\bibitem{ji2018fully}
S.~Ji, S.~Wei, and M.~Lu, ``Fully convolutional networks for multisource
  building extraction from an open aerial and satellite imagery data set,''
  \emph{IEEE Transactions on Geoscience and Remote Sensing}, vol.~57, no.~1,
  pp. 574--586, 2018.

\bibitem{City_of_NewYork}
``{NYC OpenData Building Footprints},''
  \url{https://data.cityofnewyork.us/Housing-Development/Building-Footprints/nqwf-w8eh},
  accessed: 2022-09-27.

\bibitem{dos2020regularization}
R.~C. dos Santos, M.~Galo, and A.~F. Habib, ``Regularization of building roof
  boundaries from airborne lidar data using an iterative cd-spline,''
  \emph{Remote Sensing}, vol.~12, no.~12, p. 1904, 2020.

\end{thebibliography}

\vspace{11pt}


\vfill

\end{document}